\documentclass{article} % For LaTeX2e
\usepackage{iclr2024_conference,times}
\usepackage{wrapfig}
\usepackage{amsmath,amsfonts,bm}

\def\eqref#1{equation~\ref{#1}}
\def\1{\bm{1}}

\DeclareMathAlphabet{\mathsfit}{\encodingdefault}{\sfdefault}{m}{sl}
\SetMathAlphabet{\mathsfit}{bold}{\encodingdefault}{\sfdefault}{bx}{n}

\usepackage[utf8]{inputenc} % allow utf-8 input
\usepackage[T1]{fontenc}    % use 8-bit T1 fonts
\usepackage{hyperref}       % hyperlinks
\usepackage{url}            % simple URL typesetting
\usepackage{booktabs}       % professional-quality tables
\usepackage{amsfonts}       % blackboard math symbols
\usepackage{nicefrac}       % compact symbols for 1/2, etc.
\usepackage{microtype}      % microtypography
\usepackage{xcolor}         % colors
\usepackage{amsmath}
\usepackage{amssymb}% http://ctan.org/pkg/amssymb
\usepackage{pifont}% http://ctan.org/pkg/pifont
\newcommand{\cmark}{\ding{51}}%
\newcommand{\xmark}{\ding{55}}%
\usepackage{xspace}
\usepackage{graphicx}
\usepackage{algpseudocode}
\usepackage{algorithm}
\usepackage{color}
\usepackage{tabu}
\usepackage{diagbox}
\usepackage{float}
\usepackage{placeins}
\usepackage{multirow}
\usepackage{subfigure}
\NewDocumentCommand{\yuzhen}{ mO{} }{\textcolor{red}{\textsuperscript{\textit{yuzhen}}\textsf{\textbf{\small[#1]}}}}
\newcommand{\toolName}{IceFormer\xspace}
\newcommand{\revise}[1]{\textcolor{black}{#1}}

\title{IceFormer: Accelerated Inference with Long-Sequence Transformers on CPUs}

\author{Yuzhen Mao, Martin Ester, Ke Li \\
School of Computing Science, Simon Fraser University \\
Burnaby, BC V5A 1S6, Canada \\
\texttt{$\{$yuzhenm,ester,keli$\}$@sfu.ca}
}

\iclrfinalcopy % Uncomment for camera-ready version, but NOT for submission.
\begin{document}

\maketitle

\begin{abstract}
  One limitation of existing Transformer-based models is that they cannot handle very long sequences as input since their self-attention operations exhibit quadratic time and space complexity. This problem becomes especially acute when Transformers are deployed on hardware platforms equipped only with CPUs. To address this issue, we propose a novel method for accelerating self-attention at inference time that works with pretrained Transformer models out-of-the-box without requiring retraining. We experiment using our method to accelerate various long-sequence Transformers, including a leading LLaMA 2-based LLM, on various benchmarks and demonstrate a speedup of $2.73\times - 7.63\times$ while retaining $98.6\% - 99.6\%$ of the accuracy of the original pretrained models. The code is available on our project website at~\url{https://yuzhenmao.github.io/IceFormer/}.
  %With this approach, not only can we avoid redundant computation that exists in vanilla self-attention, but we can also significantly reduce the memory footprint. Following prior work, we evaluate our model under long-sequence settings on five different tasks, which demonstrates both its effectiveness and scalability. 
\end{abstract}

\section{Introduction}

Transformers~\citep{vaswani2017attention} have powered incredible advances in NLP, as exemplified by large language models (LLMs) such as GPT-4 and LLaMA 2. Increasingly LLMs are applied to exceptionally long input sequences, which enables many exciting applications such as long-form content creation, extended conversations, and large document search and analysis~\citep{openai-gpt-4, anthropic-100k-context-windows}. While LLMs can be feasibly trained with expensive hardware accelerators (e.g. GPUs), they need to be deployed on  commodity devices, which may only be equipped with CPUs. 

However, it is currently challenging to deploy LLMs on CPUs due to their high computation cost~\citep{dice2021optimizing}. A significant computational bottleneck arises from the self-attention mechanism that is integral to Transformers -- both time and space complexity are quadratic in the sequence length. This problem is exacerbated in the context of LLMs, which are often used on very long sequences.

To handle long input sequences, there has been substantial research into reducing the quadratic time complexity of self-attention -- these methods are collectively known as \emph{efficient Transformers}. However, many do not meet the needs of LLMs and are therefore difficult to apply to LLMs. 

An ideal acceleration method for LLMs should satisfy four criteria: (1) \textbf{No retraining} -- the method should not require the model to be retrained, given the enormous computational expense of training LLMs; (2) \textbf{Generality} -- the method should be applicable to a variety of LLMs, rather than just those trained with particular constraints built-in; (3) \textbf{High accuracy} -- the method should not introduce large approximation errors, since LLMs have many attention layers and so errors from earlier layers can compound; (4) \textbf{Fast inference} -- the method should achieve fast test-time performance.

Satisfying all these criteria simultaneously is difficult, and to our knowledge no existing methods can do so. For example, Transformers with fixed attention patterns, e.g., Longformer~\citep{beltagy2020longformer}, require retraining the model before they can be used. Reformer~\citep{nikita2020reformer} requires keys to be normalized -- this requirement is not met in most pretrained models. Nystr\"{o}mformer~\citep{xiong2021nystromformer} and LARA~\citep{zheng2022linear} do not support causal masks, which are commonly found in LLMs. Low-rank methods such as Performer~\citep{choromanski2020rethinking} introduce substantial approximation errors, especially when they are not retrained/finetuned. 

\begin{figure}
  \centering
\includegraphics[width=1.0\linewidth]{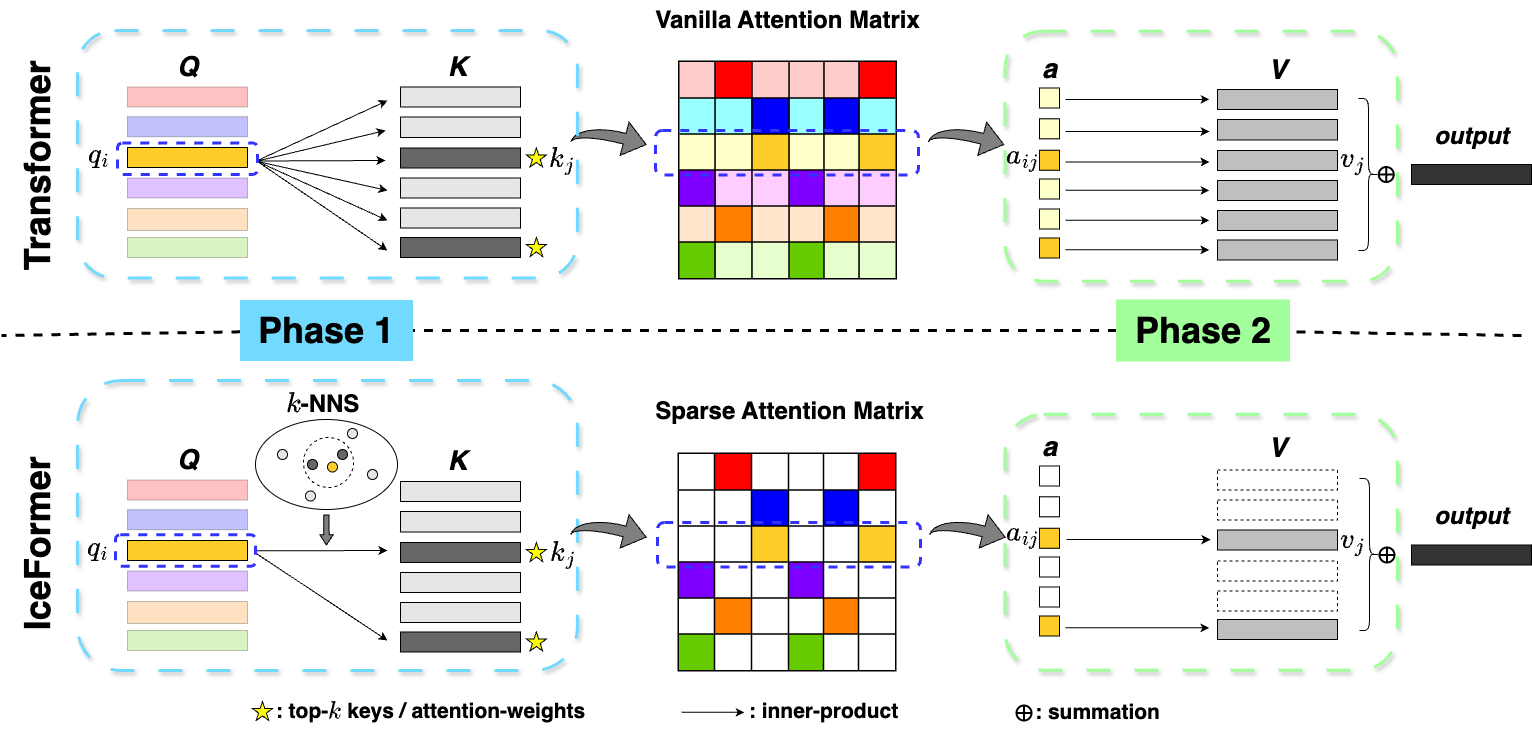}
  \caption{Comparison between Transformer~\citep{vaswani2017attention} (top row) and the proposed
method, \toolName (bottom row). We illustrate with one query and $k=2$ in $k$-NNS. In the two attention matrices presented, the top-2 largest attention weights in each row are represented by a dark color. The remaining attention weights are shown in a pale color in the vanilla attention matrix, and are set to zero (depicted in white) in the sparse attention matrix.}
  \label{fig:pipe}
\end{figure}

In this paper, we propose an acceleration method, which we dub \textbf{\toolName}  due to its ability to be applied directly in frozen models without retraining, that simultaneously satisfies the above four criteria. Specifically, \toolName (1) does not require retraining, (2) can be applied to most LLMs, (3) can approximate vanilla attention accurately, and (4) achieves significantly faster inference speeds compared to existing methods. We illustrate our method in comparison to the Transformer in Figure~\ref{fig:pipe}. As shown, the Transformer computes the attention weights $a_{ij}$ for every possible combination of query $q_i$ and key $k_j$ (Phase 1) and exhaustively enumerates all value vectors $v_j$ for each query (Phase 2). In contrast, our method takes advantage of sparsity of the attention matrix and only computes the highest attention weights and enumerates only the value vectors associated with them. %optimizes both its time and space complexity by only computing the multiplication of limited pairs of query \& key and limited pairs of attention weights \& value, utilizing $k$-Nearest Neighbor Search ($k$-NNS).

% While Reformer~\citep{nikita2020reformer} has also leveraged $k$-Nearest Neighbor Search ($k$-NNS), it imposes two constraints, namely normalization of keys and sharing of keys and queries, which are not common in vanilla Transformer models. In contrast, our approach does not impose any constraints and can work with off-the-shelf Transformer models. 

%Specifically, we approximate the standard self-attention mechanism by selectively calculating inner-products for a limited set of token pairs that exhibit high attention weights, while ignoring all other pairs. In this way, we reformulate the approximate problem as a Maximum Inner Product Search (MIPS) problem, which can be effectively addressed using a $k$-NNS algorithm~\cite{li2017fast} after applying appropriate transformation to the embeddings~\cite{keivani2017improved}. As a result, we achieve a significant reduction in both \textbf{time and space complexity}, with the level of efficiency being determined by the performance of the $k$-NNS algorithm, i.e., $O\left(d k \max \left(\log (N / k),(N / k)^{1-m / d^{\prime}}\right)+\right.$
%$\left.m k \log m\left(\max \left(\log (N / k),(N / k)^{1-1 / d^{\prime}}\right)\right)\right)$.

We conduct experiments on CPUs on the LRA~\citep{tay2020long}, ZeroSCROLLS~\citep{shaham2023zeroscrolls}, and LongEval~\citep{longchat2023} benchmarks. Across all three benchmarks, \toolName demonstrates substantially faster inference speeds than existing methods while attaining almost no accuracy loss compared to the Transformer. On the LRA benchmark, on average \toolName achieves a $7.63\times$ speedup relative to the Transformer while retaining $98.6\%$ of its accuracy. Compared to the best efficient Transformer with comparable accuracy for each task, \toolName is on average $3.04\times$ faster. On the ZeroSCROLLS benchmark, \toolName achieves a $2.73\times$ speedup on average compared to a leading LLaMA 2-based LLM while retaining $99.6\%$ of its accuracy. 
%Furthermore, our approach achieves accuracy close to that obtained using the vanilla Transformer, but also surpasses the efficiency of all baseline methods. 

% The rest of the paper is organized as follows: we review the related work in Section 2 and provide background information in Section 3. In Section 4 and 5, we present a detailed description of our proposed model. Section 6 reports the experimental results and provides analysis. Finally, we conclude the paper in Section 7.

\section{Related Work}
Efficient Transformers can be categorized along two axes: method type and retraining requirement. Along the first axis are sparsity-based methods and low-rank methods. Along the second axis are methods that can and cannot be applied to common pretrained Transformers without retraining. 

Sparsity-based methods employ a sparsified attention mechanism to capture global information and integrate it with local attention results. Some approaches aim to improve the space complexity compared to the vanilla attention mechanism without improving the time complexity, e.g., top-$k$ Attention~\citep{gupta2021memory}. Other approaches aim to improve both, e.g., Sparse Transformer~\citep{child2019generating}, Longformer~\citep{beltagy2020longformer}, and ETC~\citep{ainslie2020etc}. A substantial limitation of these models is that the tokens that are attended to are predefined and remain static, which do not adapt to varying input sequences. Because the original attention operation is permitted to attend to any token, these models must be trained with their respective predefined constraints on tokens to be attended to. Reformer~\citep{nikita2020reformer} can attend to different sets of tokens for different input sequences by using Locality Sensitive Hashing (LSH)~\citep{andoni2015practical} to group tokens into chunks and subsequently attending only to tokens within the same chunk as each query and adjacent chunks. However, Reformer imposes two constraints that are not in the original attention operation: keys must be normalized and queries and keys must be the same. Therefore, Reformer must be trained with these constraints built-in. As a result, these methods cannot be applied to pretrained, non-modified, models directly; instead, the models must be retrained with the required constraints before these methods can be used. 

Low-rank methods approximate the attention weight matrix with a low-rank matrix to reduce the quadratic time and space complexity. Examples include Linformer~\citep{wang2020linformer} and Performer~\citep{choromanski2020rethinking}, which decompose the attention weight matrix into a product of tall and wide matrices consisting of learned linear features or random features of the keys and queries, respectively. However, these Transformers typically introduce significant approximation errors because attention weight matrices produced by the original attention operation, especially in the case of long input sequences, typically have high rank. Consequently, models that use these approaches must be trained with low-rank approximations built-in, in order to learn to be robust to the associated approximation errors. As a result, these approaches cannot be applied to pretrained, non-modified, models directly; instead, the models must be retrained with the required approximations before these methods can be used. 
Other approaches provide more general methodologies that can leverage weights pretrained with standard Transformers without retraining. These Transformers accelerate the execution of the standard attention operation without altering the underlying architecture. Two examples are Nystr\"{o}mformer~\citep{xiong2021nystromformer} and LARA~\citep{zheng2022linear}, which replace the softmax structure in the self-attention mechanism with the product of separately activated query and key matrices. Nystr\"{o}mformer utilizes the Nystr\"{o}m method, while LARA combines randomized attention (RA) and random feature attentions (RFA)~\citep{peng2021random} to reconstruct the attention weight matrix. In another example, H-Transformer-1D~\citep{zhu-soricut-2021-h} recursively divides the attention weight matrix into blocks and truncates the small singular values of each off-diagonal blocks. All these approaches leverage low-rank approximations, as opposed to sparsity. 

Other works propose hardware-specific optimizations without aiming to improve the computational complexity. Examples include FlashAttention~\citep{dao2022flashattention}, which optimizes reads and writes between levels of GPU memory, and H2O~\citep{zhang2023h}, which dynamically retains a balance of recent and heavy hitters tokens by a KV cache eviction policy. These strategies are dependent on implementation and are specific to particular hardware platforms (e.g. GPU).

\section{Notation and Preliminaries
}

Mathematically, the attention operation takes three matrices as input, $\mathbf{K} \in \mathbb{R}^{m \times d}, \mathbf{Q} \in \mathbb{R}^{n \times d}, \mathbf{V} \in \mathbb{R}^{m \times d'}$, which denote keys, queries and values respectively, and outputs a matrix $\mathbf{O} \in \mathbb{R}^{n \times d'}$. Optionally, it may also take in a mask as input, $\mathbf{S} \in \mathbb{R}^{n \times m}$, whose entries are either 0 or 1. The $i$th rows of $\mathbf{K}$, $\mathbf{Q}$, $\mathbf{V}$ and $\mathbf{O}$, denoted as $\mathbf{k}_i$, $\mathbf{q}_i$, $\mathbf{v}_i$ and $\mathbf{o}_i$, represent the $i$th key, query, value and output respectively. The entry of $\mathbf{S}$ in the $i$th row and $j$th column, denoted as $s_{i,j}$, represents whether the $i$th query is allowed to attend to the $j$th key --- if it is $1$, it would be allowed; if it is $0$, it would not be. A common masking scheme is the causal mask, where $s_{i,j}$ is $1$ if $i \geq j$ and $0$ otherwise. Keys and queries have the same dimension $d$, and each key is associated with a value, and so the number of keys and values is the same and denoted as $m$. 

First the attention operation computes the attention weight matrix $\mathbf{A} \in \mathbb{R}^{n \times m}$. Its entry in the $i$th row and $j$th column, denoted as $a_{i,j}$, is computed with the following formula:

\begin{equation}\label{eq:attn_weights}
a_{i,j}=\frac{s_{i,j} \exp \left( \frac{\mathbf{q}_i^\top \mathbf{k}_{j}}{\sqrt{d}} \right)}{\sum_{j'=1}^{m} s_{i,j'} \exp \left( \frac{\mathbf{q}_i^\top \mathbf{k}_{j'}}{\sqrt{d}} \right)}
\end{equation}

Then the attention operation combines the values with the attention weights in the following way:
\begin{equation}\label{eq:value_comb}
\mathbf{o}_i = \sum_{j=1}^{m} a_{i,j} \mathbf{v}_j
\end{equation}

The attention matrix $\mathbf{A}$ is typically sparse~\citep{nikita2020reformer, gupta2021memory}, i.e., in each row of $\mathbf{A}$, only a few attention weights have significant (large) values, while the majority of the remaining values are close to zero. Suppose we can somehow identify the $k$ unmasked keys that receive the highest attention weights for each query $\mathbf{q}_i$ without computing the attention weights for all keys. Then, the original attention matrix $\mathbf{A}$ can be approximated by only computing the inner product for the identified keys, which can save significant amount of time and computational resource.

\section{IceFormer: Accelerated Self-Attention for General Keys without Retraining}

% \subsection{Reduction to $k$-Nearest Neighbour Search for Normalized Keys\label{sec:red_norm_key}}

%\subsection{Reduction to $k$-Nearest Neighbour Search\label{sec:red_norm_key}}

%The observation that the attention matrix of the self-attention mechanism is sparse leads us to a natural follow-up inquiry: \textit{how can we effectively identify the pairs of tokens that exhibit significant (large) values without computing the whole attention matrix?} 

To build a general-purpose retraining-free acceleration method, our approach must not require modifications to the attention mechanism to change attention patterns or the introduction of new model parameters to capture regularities in the attention patterns. This precludes popular strategies such as attention mechanisms with predefined sparse attention patterns, e.g., \citep{child2019generating,beltagy2020longformer,ainslie2020etc}, and learned dimensionality reduction of keys and queries, e.g., \citep{wang2020linformer,choromanski2020rethinking}. 

Consequently, it is difficult to design an acceleration method that exploits known regularities in the attention patterns without imposing the retraining requirement. We therefore aim to design an acceleration method that does not make assumptions on the existence of regularity in the attention patterns. In order to improve on the $O(mn)$ complexity of vanilla attention, we need to adaptively identify the most important keys (i.e., those that receive the highest attention weights) without computing all attention weights. This seems like a chicken-and-egg problem: how can we know which attention weights are highest without comparing them to all the other attention weights?

Remarkably, in the special case of normalized keys, as proposed in~\citet{nikita2020reformer}, this can be done by leveraging $k$-nearest neighbour search ($k$-NNS) to identify the $k$ most important keys for each query. This relies on the following mathematical fact, whose derivation is in included in Sect.~\ref{proof1} of the appendix: if $\Vert \mathbf{k}_{j} \Vert_2 = 1$ for all $j$, $\arg\max_j a_{i,j} = \arg\min_j \Vert \mathbf{q}_i - \mathbf{k}_{j} \Vert_2^2$. 

However, this fact only holds when all the keys have the same norm -- it is not true when different keys differ in their norms. Intuitively, this is because the norms of keys can modulate the attention weights they receive, all else being equal. So if key A has a larger norm than key B, key A can receive a higher attention weight than key B even if key A is farther from the query than key B. As a result, na\"{i}vely applying $k$-NNS in the general case would fail to identify the most important keys.

In this paper, we develop an acceleration method that does not require retraining or impose any constraints on keys. It is both accurate and computationally efficient, and can also work with attention masks that are common in Transformers, such as causal masks. Below we will describe the details.

\subsection{General Retraining-Free Accelerated Attention}\label{sec:math}

%We propose a method for accelerating attention that can be applied directly to vanilla attention mechanisms without constraints on keys without retraining. We will reduce the general problem of accelerating attention with unconstrained keys to the problem of accelerating attention with normalized keys, and then solve the latter. 

Instead of applying $k$-NNS to the original keys directly, we will first embed the keys and queries into a higher dimensional space. Inspired by \citet{neyshabur2015symmetric}, we choose the following key and query embedding functions, which we denote as $T_K: \mathbb{R}^d \rightarrow \mathbb{R}^{d+1}$ and $T_Q: \mathbb{R}^d \rightarrow \mathbb{R}^{d+1}$:

%In the self-attention mechanism setting, $S$ corresponds to the set of key vectors, and $\vec{q}$ corresponds to a query vector of the Transformer. Formally, given the index list $l_{i}$ of the selected top-$k$ keys for the query $\mathbf{Q}_{i}$, reduced self-attention mechanism can be formulated as follows. 

%Notably, it has been proved that MIPS can be solved as a $k$-NNS problem~\cite{keivani2017improved}, which is much easier to be optimized and approximated.

%We first review the connection between MIPS and $k$-NNS as follows. 

%Given the previous setting of Eq.~\ref{eq:4}, 

%after applying transformation $P: \mathbb{R}^d \rightarrow \mathbb{R}^{d+1}$ to points in $S$ and applying transformation $Q: \mathbb{R}^d \rightarrow \mathbb{R}^{d+1}$ to query $\vec{q}$, 
\begin{align}
T_K(\mathbf{k}_j) & =\begin{bmatrix}\mathbf{k}_j / c & 
\sqrt{1-\|\mathbf{k}_j\|_2^2 / c^2}\end{bmatrix}^\top \\
T_Q(\mathbf{q}_i) & =\begin{bmatrix}
\mathbf{q}_i / \|\mathbf{q}_i\|_2 &
0
\end{bmatrix}^\top
\end{align}
where $c \geq \max _{j'}\|\mathbf{k}_{j'}\|_2$ is at least the maximum norm across all keys. 

%It is easy to see that $\Vert T_K(\mathbf{k}_j) \Vert_2 = 1$ and $\Vert T_Q(\mathbf{q}_i) \Vert_2 = 1$. So both the transformed keys and queries are normalized, which would then allow us to apply the method covered in Sect.~\ref{sec:red_norm_key}. 

It turns out that the $k$ most important keys can be identified by performing $k$-NNS on the key embeddings using the query embedding. We will show this below:
\begin{align}
\arg\max_j a_{i,j} & = \arg\max_j \mathrm{softmax}_j\left( \left\{\frac{\mathbf{q}_i^\top \mathbf{k}_{j'}}{\sqrt{d}}\right\}_{j'=1}^{m} \right)\hphantom{ c\Vert \mathbf{q}_i \Vert_2 + \mathbf{k}_{j}^\top \mathbf{k}_{j} / c^2 + 1 - \|\mathbf{k}_{j}\|_2^2 / c^2}
\end{align}

\begin{align}
& = \arg\max_j \frac{\mathbf{q}_i^\top \mathbf{k}_{j}}{\sqrt{d}} \\
%& = \arg\max_j \mathbf{q}_i^\top \mathbf{k}_{j} \\
& = \arg\min_j 1 -  2\mathbf{q}_i^\top\mathbf{k}_{j} / c\Vert \mathbf{q}_i \Vert_2 + 1 \\
& = \arg\min_j \mathbf{q}_i^\top \mathbf{q}_i / \Vert \mathbf{q}_i \Vert_2^2 -  2\mathbf{q}_i^\top\mathbf{k}_{j} / c\Vert \mathbf{q}_i \Vert_2 + \mathbf{k}_{j}^\top \mathbf{k}_{j} / c^2 + 1 - \|\mathbf{k}_{j}\|_2^2 / c^2 \\
& = \arg\min_j \Vert \mathbf{q}_i / \Vert \mathbf{q}_i\Vert_2 - \mathbf{k}_{j} / c \Vert_2^2 + 1 - \|\mathbf{k}_{j}\|_2^2 / c^2 \\
& = \arg\min_j \Vert T_Q(\mathbf{q}_i) - T_K(\mathbf{k}_{j}) \Vert_2^2
\end{align}

%solving MIPS in the original space is equivalent to solving $k$-NNS in the transformed space and the following equation holds:
%$$
%\arg \max _{\vec{x} \in S} \vec{q}^{\top} \vec{x}=\arg \min _{\vec{x} \in S}\left\|P(\vec{x})-Q(\vec{q})\right\|_2
%$$
%where $P$ and $Q$ are defined as follows~\cite{keivani2017improved}:
%$$
%P(\vec{x})=\left(\frac{\vec{x}}{\beta}, \sqrt{1-\frac{\|\vec{x}\|_2^2}{\beta^2}}\right), Q(\vec{x})=\left(\frac{\vec{x}}{\|\vec{x}\|_2}, 0\right)
%$$
%with $\beta=\max _{\vec{x} \in S}\|\vec{x}\|_2$ is the maximum norm among all data points in $S$.

\subsection{Accurate $k$-NNS for Accelerated Attention}

The problem of $k$-NNS is one of the most well studied problems in theoretical computer science. Many algorithms have been developed, and often significant speedups can be obtained by allowing for mistakes with some probability. Such algorithms are known as randomized algorithms. 

In the context of LLMs, the number of attention layers is typically high and so errors from earlier layers can compound. Therefore, it is essential for the $k$-NNS algorithm to achieve high accuracy. Choosing an appropriate $k$-NNS algorithm is therefore crucial. 

%, because the algorithmic properties of accelerated attention will be inherited from the $k$-NNS algorithm. %There are two major aspects when selecting the $k$-NNS algorithms. (1) High accuracy of approximation.  

%For the first aspect - high accuracy of approximation, we will consider two axes in the space of $k$-NNS algorithms that are most relevant: bucketing-based vs. ranking-based algorithm, and exact vs. approximate algorithm. 

%Bucketing vs. ranking
%Exact vs. approximate

\begin{figure}[htb]
\centering
\includegraphics[width=0.93\linewidth]{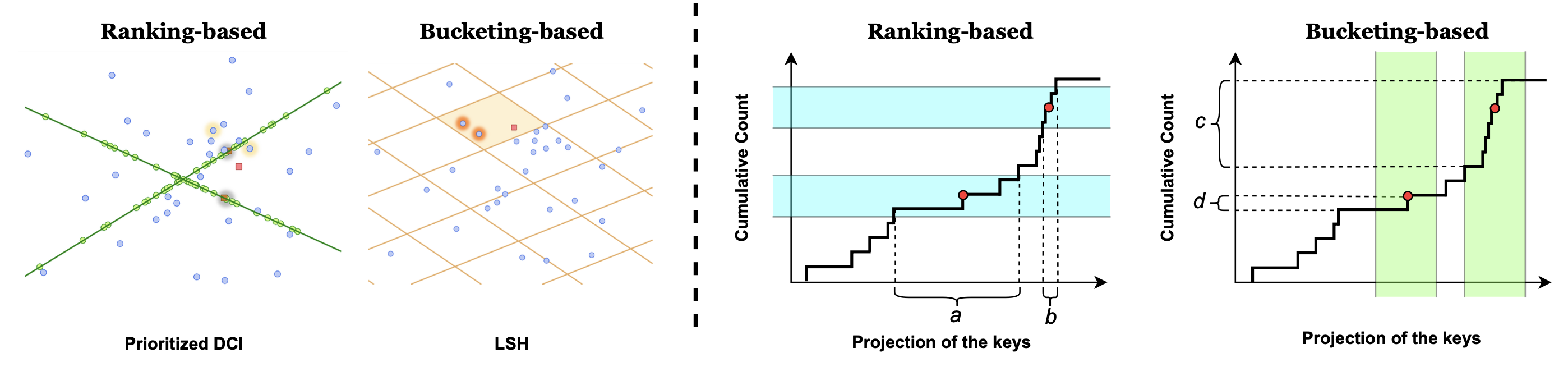}
  \caption{Difference between ranking-based and bucketing-based $k$-NNS. Left: illustration of two $k$-NNS methods, Prioritized DCI (ranking-based) and LSH (bucketing-based). Right: the number of keys whose projections are less than a threshold. Ranking-based algorithms return a fixed number of keys are most similar to the query under projection (shown as a fixed-size row), which effectively filters out points outside a variable-sized window on the projections. Bucketing-based algorithms use a fixed-size window (shown as a fixed-size column) and return all keys whose projections lie within it. }
  \label{fig:lsh}
\end{figure}

Most $k$-NNS algorithms are bucketing-based, which places keys into discrete buckets and searches over buckets that contain the query. On the other hand, ranking-based algorithms compares the rankings of different keys relative to the query and searches over highly ranked keys. A bucketing-based algorithm effectively uses a fixed threshold on similarity, and so a variable number (including zero) of keys can meet the threshold; on the other hand, a ranking-based algorithm returns a fixed number of keys, which effectively amounts to choosing a variable threshold on similarity based on the distribution of keys, as shown in Figure~\ref{fig:lsh}. An example of a bucketing-based algorithm is locality-sensitive hashing (LSH)~\citep{indyk1998approximate}, and an example of a ranking-based algorithm is Prioritized DCI~\citep{li2017fast}. As shown in Figure~\ref{fig:lsh}, LSH hashes each key into a bucket associated with the hash value, whereas Prioritized DCI ranks keys along random directions. 

%Each hash bin is LSH can be viewed as a bucket (as shown in Figure~\ref{fig:lsh}). As a result, the number of keys returned by LSH may be variable and could even be zero for some queries whose similarity to all keys fall below a threshold.
%One example of a ranking-based algorithm is Prioritized DCI~\citep{li2017fast}, which ranks points along random projection directions rather than bucketing them (as shown in Figure~\ref{fig:lsh}). Regardless of how similar/dissimilar the keys to the query, it returns a fixed number of keys, which is equivalent to choose a variable threshold dynamically, dependent on the current query and its similarity to the keys. 

For accelerating attention, we posit that ranking-based algorithms are better suited than bucketing-based algorithms, because attention weights depend on how different keys compare to one another, rather than an absolute evaluation of each key against a fixed threshold. Therefore, ranking-based algorithms can yield better recall of truly important keys. 

%, and so we expect the latter to yield a better approximation to attention. 

%Therefore, after transforming the problem, the fundamental bottleneck becomes the effectiveness of the $k$-NNS algorithm. Locality-Sensitive Hashing (LSH)~\cite{har2012approximate} is often regarded as a $k$-NNS algorithm, sharing certain similarities with DCI. However, it is crucial to differentiate between LSH and DCI, as they possess distinct properties that influence the construction of the reduced attention matrix (Eq.~\ref{eq:6}). These differences ultimately lead to variations in the overall performance of the model.

\subsection{Fast $k$-NNS for Accelerated Attention}

In a Transformer, the keys in an attention layer depend on the output from the preceding attention layer. Therefore, a database needs to be constructed for each attention layer. Therefore, it is important to choose a $k$-NN algorithm that attains both fast construction and querying. 

%When adapting $k$-NNS algorithm to the Transformer, it is necessary to construct a different database for each attention-layer before querying the top-$k$ closest keys for queries. Therefore, both the construction time and query time of a $k$-NNS algorithm need to be taken into the consideration when evaluating its latency.

Moreover, in the context of LLMs, many popular models use decoder-only architectures. The attention layers in such architectures use causal masks to prevent the currently generated token to depend on future yet-to-be-generated tokens. Such masked attention is equivalent to excluding the masked out keys from the set of keys the $k$-NNS algorithm operates over. So each time a token is generated, one key becomes unmasked. Instead of constructing a new database each time a token is generated, it is more efficient to add keys incrementally to the database for $k$-NNS. 

Fortunately, Prioritized DCI is efficient at both the construction and querying stages. If the number of random projection directions $p$ is nearly as large as the intrinsic dimensionality of the data $d' \geq 1$ and the number of nearest neighbours $k$ to look for is small, Prioritized DCI can return the exact $k$-nearest neighbours for a query with high probability within approximately $\tilde{O}(d k^{p / \tilde{d}} m^{1-p / \tilde{d}})$ time, where $\tilde{O}(\cdot)$ suppresses log factors. Its preprocessing is lightweight, and so only needs $O(dpm)$ time. If we compare this to the computational complexity of vanilla attention of $O(dmn)$, observe that there is no longer a term that depends on $mn$, and so there is no longer the quadratic dependence on sequence length. Later in section~\ref{sec:knn}, we also empirically validate the efficiency of Prioritized DCI and found it to be faster than \revise{eleven} other leading $k$-NNS algorithms. 

%by comparing it with nine other $k$-nearest neighbour search algorithms on fashion-mnist-784 dataset, which supports our second aspect (Inference speed) for choosing a suitable $k$-NNS algorithm for the accelerating attention.

To support causal masking, we extended the implementation of Prioritized DCI to support incremental database updates. This can be done efficiently, since the data structure consists of sorted lists, so insertions and deletions can be done in $O(\log m)$ time if they are implemented as binary search trees.

\section{Experiments}

\begin{wrapfigure}{r}{0.39\textwidth}
  \centering
  \includegraphics[width=1\linewidth]{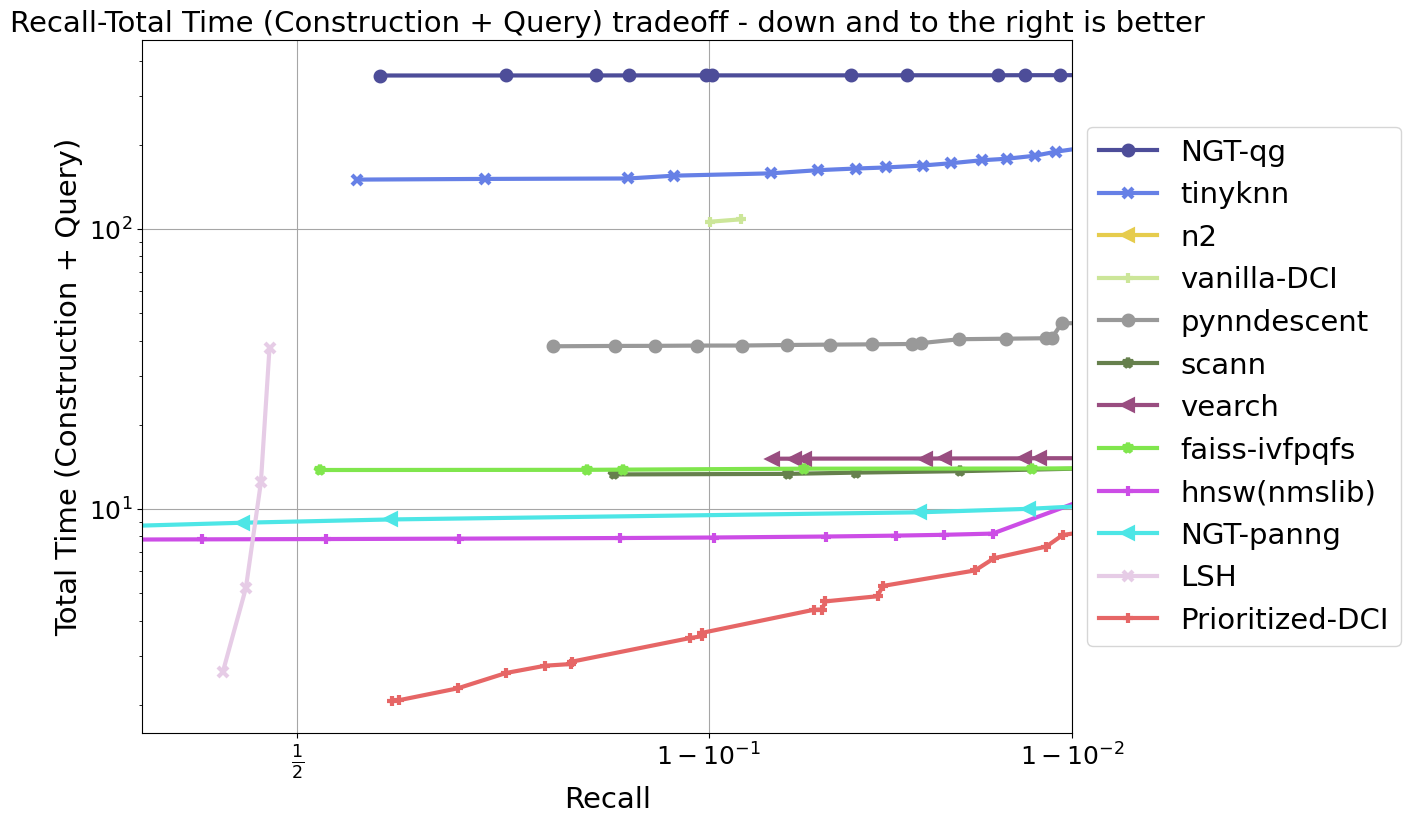}
  \caption{Comparison between \revise{twelve} $k$-NNS algorithms on fashion-mnist-784 dataset. There are in total 60,000 keys and 10,000 queries with 784 dimensions. The task is to find top-10 closest neighbours from the entire set of keys for every query. X-axis: Average recall across all the queries; Y-axis: Total latency (seconds) including database construction and querying.}
  \label{fig:knns}
\end{wrapfigure}

In this section, we will compare the recall-latency trade-off between different $k$-NNS algorithms and then analyze the performance of \toolName on the LRA benchmark~\citep{tay2020long}, which is a popular benchmark for long-context Transformers~\citep{zhu-soricut-2021-h, xiong2021nystromformer, zheng2022linear}. Next we will demonstrate the advantages of \toolName applied to LLMs with long prompts as input on the ZeroSCROLLS benchmark~\citep{shaham2023zeroscrolls} and the LongEval benchmark~\citep{longchat2023}. To ensure robustness of results, we used a variety of CPUs for our experiments -- we used Intel(R) Core(TM) i7-6850K 6-Core for the LRA experiments, AMD Ryzen 9 5950X 16-Core for the ZeroSCROLLS experiments, and AMD Ryzen 9 5900X 12-Core for the LongEval experiments.

% The goal is to show that under the long-sequence settings, our proposed model could speed up the vanilla self-attention significantly (\textbf{up to 10 times faster}) and maintain a relatively good model accuracy at the same time.
\subsection{Different $k$-NNS algorithms comparison}\label{sec:knn}
We compare the recall of true nearest neighbours and total construction and querying time of 12 $k$-NNS algorithms, including Prioritized DCI and the best performing algorithms from ANN benchmarks~\citep{aumuller2017ann}, on the Fashion MNIST dataset in Figure~\ref{fig:knns}. As shown, Prioritized DCI achieves the best recall-latency trade-off compared to other algorithms, which demonstrates its suitability in our setting, which requires fast construction and querying.

\subsection{Evaluation on Long Range Arena (LRA) Benchmark}
\paragraph{Datasets and Metrics.}
LRA consists of five different tasks: ListOps~\citep{nangia2018listops}, document retrieval (Retrieval)~\citep{radev2013acl}, text classification (Text)~\citep{maas-EtAl:2011:ACL-HLT2011}, CIFAR-10 image classification (Image)~\citep{krizhevsky2009learning} and Pathfinder~\citep{linsley2018learning}.  Specifically, all the five tasks consist of sequences with at most 4k tokens. We summarize the dataset information in the appendix~\ref{sec:lra_data} for more details. In this experiment, we follow the train/test splits from~\citet{tay2020long} and report the test dataset classification accuracy, average running time of the attention module, and CPU memory usage during inference for each task.

\paragraph{Baselines.}
In addition to the vanilla Transformer, we compare with Nystr\"{o}mformer~\citep{xiong2021nystromformer}, H-Transformer-1D~\citep{zhu-soricut-2021-h}, LARA~\citep{zheng2022linear}, Reformer~\citep{nikita2020reformer}, \revise{Longformer~\citep{beltagy2020longformer}, Performer~\citep{choromanski2020rethinking}, and Linformer~\citep{wang2020linformer}}. In order to compare with Reformer, we train a Transformer model with shared $\mathbf{Q}$ and $\mathbf{K}$ according to~\citet{nikita2020reformer}. \revise{For Longformer and Linformer, as they introduce additional parameters, we randomly initialize these parameters when loading the pre-trained weight from the vanilla Transformer.} For fair comparisons, we use
the LRA evaluation benchmark implemented in PyTorch by~\citep{xiong2021nystromformer}, and only replace the self-attention module while making other parts of each model exactly the same as the vanilla Transformer. 

% For Nystromformer, H-Transformer-1D and LARA, each has one parameter that can be used to tune the model sparsity; while for Reformer, there are two such parameters.

\paragraph{Implementation Details.}
For each task, we begin by training a base model using GPU with a vanilla Transformer architecture. Then we replace the vanilla attention module with one of the eight efficient attention modules mentioned earlier and directly apply the pre-trained weights for inference.  To ensure fair comparison, we adjust the batch size to 1, eliminating the need for a padding mask since our proposed \toolName automatically ignores padding masks during inference. Note that because of the additional shared-KQ constraint, for the Pathfinder task, our attempts to train a shared-KQ Transformer were unsuccessful. As a result, we have excluded the corresponding results from the subsequent analysis. Additionally, during the inference, we utilize a total of 4 CPU threads. For more comprehensive details, please refer to the appendix~\ref{sec:setting}.

\paragraph{Inference Results.}
Ideally, the accuracy of the vanilla Transformer (non-shared-KQ) serves as an upper bound for the approximated accuracy of the other seven models (\toolName (non-shared-KQ), Nystr\"{o}mformer, H-Transformer-1D, LARA, \revise{Longformer, Performer, and Linformer}). Similar for the shared-KQ Transformer. Also, the attention module inference time of the vanilla Transformer would be the longest, with other efficient Transformers achieving shorter inference times at the cost of sacrificing prediction accuracy. Table~\ref{tab:2} presents the prediction accuracy and inference time of the attention module for each method. The hyper-parameter settings are listed in the appendix~\ref{sec:hyper}. In general, our proposed \toolName consistently outperforms all efficient Transformers, offering the best accuracy approximation while requiring the least inference time across all five tasks. This demonstrates the generalizability and effectiveness of our model.

\begin{table}[htb]
\caption{\revise{The performance of vanilla Transformer, and eight approximate attention methods on the LRA benchmarks.}}
\centering
\scalebox{0.621} {
\setlength{\tabcolsep}{4pt}
\begin{tabular}{c|c|cc|cc|cc|cc|cc}
\toprule \multirow{2}{*}{Method}  & \multirow{2}{*}{shared-KQ}   &  \multicolumn{2}{c|}{ListOps}  &  \multicolumn{2}{c|}{Text} & \multicolumn{2}{c|}{Retrieval} & \multicolumn{2}{c|}{Image} &  \multicolumn{2}{c}{Pathfinder}   \\
\cline { 3 - 12 }
& &  Acc & Time (s) & Acc & Time (s) & Acc & Time (s) & Acc &  Time (s) & Acc & Time (s) \\
\midrule
\midrule
\multirow{2}{*}{Transformer~\citep{vaswani2017attention}}  
& \xmark & 0.4255 & 2.9208 & 0.6019 & 0.6933 & 0.6586 & 8.3588 & 0.4132 & 4.9303 & 0.7514 & 0.9620 \\ 
& \cmark  & 0.4145 & 2.9134  & 0.5986 &  0.6603 & 0.6681 & 6.7946 & 0.3844 & 5.9804 & / & / \\ 
\midrule
\midrule
{Reformer~\citep{nikita2020reformer}} & \cmark & 0.4121 &  1.4281 &  0.5941 & 0.2288  & 0.6467 & 1.4751 & 0.3726 & 3.6927 & / & / \\
{LARA~\citep{zheng2022linear}} & \xmark  &  0.4125 &  0.6146 & 0.5831 & 0.2348 & 0.6401 & 1.8605 & 0.3094 & 2.6720 & 0.7380 & 0.5961 \\
{Nystr\"{o}mformer~\citep{xiong2021nystromformer}} & \xmark & 0.4128 & 0.7994 & 0.5838 & 0.3542 & 0.6540 & 2.4179 &  0.3754 & 1.7644 & 0.7176 & 0.9927 \\
{H-Transformer-1D~\citep{zhu-soricut-2021-h}} & \xmark & 0.3265 & 1.9301 &  0.5944 & 0.4811 & 0.5808 & 3.5605 & 0.2286 & 1.2586 & 0.5286 & 0.5708 \\
\revise{{Longformer~\citep{beltagy2020longformer}}} & \revise{\xmark} & 
\revise{0.1975} & 
\revise{0.7406} &  
\revise{0.5236} & 
\revise{0.9862} & \revise{0.4918} & 
\revise{1.0443} & 
\revise{0.1488} & 
\revise{0.5451} & \revise{0.5009} & 
\revise{0.5899} \\
\revise{{Performer~\citep{choromanski2020rethinking}}} & \revise{\xmark} & \revise{0.1975} & 
\revise{0.6571} &  
\revise{0.5000} & 
\revise{0.3327} & 
\revise{0.4974} & 
\revise{1.2058} & \revise{0.1345} & 
\revise{0.6404} & \revise{0.5056} & 
\revise{0.6395} \\
\revise{{Linformer~\citep{wang2020linformer}}} & \revise{\xmark} & \revise{0.1975} & \revise{3.1532} & 
\revise{0.5088} & \revise{1.8912} & \revise{0.4940} & \revise{1.6878} & \revise{0.1064} & \revise{0.7387} & \revise{0.5022} & \revise{1.3141} \\
\midrule
\midrule
\multirow{2}{*}{\toolName (ours)}  
& \xmark  & \textbf{0.4153} & \textbf{0.3766} & \textbf{0.5978} & \textbf{0.0921} & \textbf{0.6541} & \textbf{0.8337} & \textbf{0.4046} & \textbf{0.5076} & \textbf{0.7442} & \textbf{0.3058} \\
& \cmark & \textbf{0.4124} & \textbf{0.4678}  &  \textbf{0.6001} & \textbf{0.0903} &  \textbf{0.6602} & \textbf{0.8480} & \textbf{0.3752} & \textbf{0.9581} & / & /  \\
\bottomrule
\end{tabular}
}
\label{tab:2}
\end{table}

\paragraph{Speed \& Accuracy Trade-off.}

\begin{figure}
  \centering
\includegraphics[width=1\linewidth]{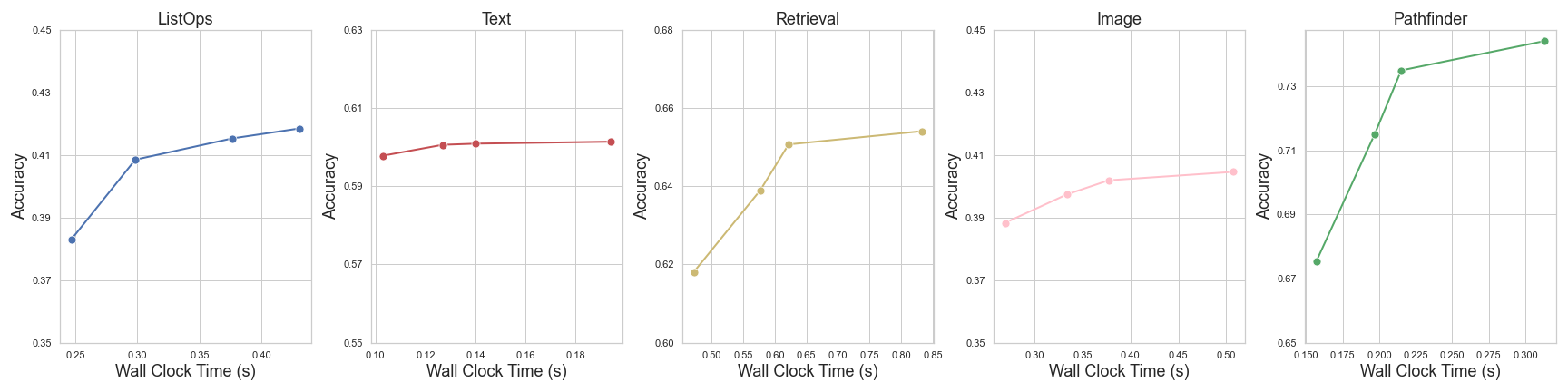}
  \caption{Tradeoff between speed and accuracy as $k$ varies on five LRA tasks. The horizontal axis of each plot is the averaged wall clock time of attention module, and the vertical axis is the model prediction accuracy. Each point corresponds to a value of $k$ in the following set: \{3, 5, 8, 10\}.}
  \label{fig:tradeoff}
\end{figure}

For \toolName, increasing the extent of approximation generally improves model efficiency but can lead to a decrease in prediction performance. Here, we study how the extent of approximation affects inference speed and accuracy by varying the number of returned candidates of \toolName, $k$, from 3 to 10 for each task and present the results in Figure~\ref{fig:tradeoff}. From the figure, we observe that across all tasks, when $k$ becomes larger, \toolName achieves improved prediction accuracy but becomes less efficient.  

\paragraph{Memory Complexity Analysis.}
Table~\ref{tab:4} summarizes the maximum memory usage for each method during inference. We employ the same hyper-parameters as in Table~\ref{tab:2} and maintain a batch size of 1 to eliminate the need for padding masks. The table reveals that \toolName consistently exhibits the lowest peak memory usage across all tasks. In comparison to the vanilla Transformer, \toolName achieves memory savings of up to 0.862 GB.
\begin{table}[H]
\caption{Peak memory usage (GB) on LRA benchmark. The peak memory usage is the total memory usage of the whole program, which includes the memory for the Prioritized DCI database/index.}
\centering
\scalebox{0.621} {
\setlength{\tabcolsep}{4pt}
\begin{tabular}{c|c|c|c|c|c|c}
\toprule {Method}  & {shared-KQ}   &  {ListOps}  &  {Text} & {Retrieval} & {Image} & {Pathfinder}  \\
\midrule
\midrule
\multirow{2}{*}{Transformer~\citep{vaswani2017attention}}  
& \xmark & 3.729 & 4.327 & 5.031 & 3.778 & 3.926 \\ 
& \cmark  & 3.631 & 4.265 & 4.877 & 3.740 & / \\
\midrule
\midrule
{Reformer~\citep{nikita2020reformer}} & \cmark & 3.623 & 3.983 & 4.250 & 3.687 & / \\
{LARA~\citep{zheng2022linear}} & \xmark   & 3.584 & 4.129 & 4.566 & 3.772 & 3.943\\
{Nystr\"{o}mformer~\citep{xiong2021nystromformer}} & \xmark & 3.478 & 3.982  & 4.375 & 3.463 & 3.845 \\
{H-Transformer-1D~\citep{zhu-soricut-2021-h}} & \xmark & 3.883 & 4.328 & 4.543 & 3.553 & 3.603 \\
\midrule
\midrule
\multirow{2}{*}{\toolName (ours)}  
& \xmark  & \textbf{3.374 }& \textbf{3.834} &\textbf{ 4.169} & \textbf{3.304} & \textbf{3.465} \\
& \cmark & \textbf{3.306} & \textbf{3.756} & \textbf{4.053} & \textbf{3.286} & / \\
\bottomrule
\end{tabular}
}
\label{tab:4}
\end{table}

\subsection{Evaluation on Large Language Model (LLM)}\label{sec:llm-exp}
We evaluate \toolName in the LLM setting as well. Specifically, we utilize \toolName to accelerate the prompt processing process in LLMs. We pick Vicuna-7b-v1.5-16k~\citep{zheng2023judging}, which is fine-tuned from LLaMA 2~\citep{touvron2023llama2} and is one of the top-performing open-source LLMs with a context length up to 16K tokens, for the following experiment. For more comprehensive details including the choice of $k$ in $k$-NNS of \toolName, please refer to the appendix~\ref{sec:llm}.

For the following LLM experiments, we do not compare \toolName with Reformer, LARA and Nystr\"{o}mformer for the following reasons: Reformer requires keys and queries to be shared, which is not the case in pre-trained LLMs; Longformer only proposed a way to speed up
the encoder part of the Transformer, thus cannot be applied to decoder-only LLMs; LARA and Nystr\"{o}mformer group different tokens into different clusters and so cannot handle causal masks in LLMs, which use decoder-only architectures. All baselines that require retraining (Longformer, Performer and Linformer) are also excluded from the comparison. More details can be found in the appendix~\ref{appendix:landmark}.

\paragraph{ZeroSCROLLS Results.}
We compare \toolName with \revise{the vanilla Vicuna-7b-v1.5-16k model} and H-Transformer-1D \revise{applied to Vicuna-7b-v1.5-16k} on the ZeroSCROLLS benchmark~\citep{shaham2023zeroscrolls} which is specifically designed for LLMs and contains ten diverse natural language tasks that require understanding long input contexts, including summarization, question answering, aggregated sentiment classification and information reordering. Each task has a different sequence length varying between 3k and 10k. We measure ZeroSCROLLS scores and latency of the attention module. Table~\ref{tab:zeroscroll} shows that \toolName achieves up to 3.0$\times$ speed-up compared to standard self-attention while attaining at least 99.0\% of the vanilla unaccelerated model performance at the same time. 

\begin{table}[h]
  \caption{The performance of \revise{the vanilla Vicuna-7b-v1.5-16k model}, H-Transformer-1D and \toolName on the ZeroSCROLLS benchmarks. Numbers in parentheses indicate the relative comparison to the vanilla unaccelerated model, denoted as \revise{Vicuna-7b-v1.5-16k}. We employ the same abbreviations for metric and task names as specified in the original paper~\citep{shaham2023zeroscrolls}. We refer interested readers to the original paper for the details.}
  \centering
  \scalebox{0.621} {
  \begin{tabular}{lc|c|c|c}
    \toprule
    {Task (\#tokens)} & {Metric} & \revise{Vicuna-7b-v1.5-16k} & H-Transformer-1D  &\textbf{\toolName} \\
    \midrule
    \midrule
    \multirow{2}{*}{GvRp (8k)} & $R_{geo} \uparrow$ & 11.0 (100\%) & 6.8 (61.8\%) & \textbf{11.0 (100\%)} \\
    & Time (s) & 5.07 (1.0$\times$) & 4.22 (1.2$\times$) &  \textbf{1.89 (2.7$\times$)} \\
    \midrule
    \multirow{2}{*}{SSFD (8k)} & $R_{geo} \uparrow$ & 13.5 (100\%) & 6.3 (46.7\%) & \textbf{13.5 (100\%)} \\
    & Time (s) & 5.02 (1.0$\times$) & 4.18 (1.2$\times$) & \textbf{1.81 (2.8$\times$)} \\
    \midrule
    \multirow{2}{*}{QMsm (9k)} & $R_{geo} \uparrow$ & 16.9 (100\%) & 10.7 (63.3\%) & \textbf{16.8 (99.4\%)} \\
    & Time (s) & 6.47 (1.0$\times$) & 4.62 (1.4$\times$) & \textbf{2.51 (2.6$\times$)} \\
    \midrule
    \multirow{2}{*}{SQAL (8k)} & $R_{geo} \uparrow$ & 18.9 (100\%) & 7.3 (38.6\%) & \textbf{18.9 (100\%)} \\
    & Time (s) & 5.01 (1.0$\times$) & 2.27 (2.2$\times$) & \textbf{1.92 (2.6$\times$)} \\
    \midrule
    \multirow{2}{*}{Qspr (5k)} & F1 $\uparrow$& 34.2 (100\%) & 6.2 (18.1\%) & \textbf{34.0 (99.4\%)} \\
    & Time (s) & 2.03 (1.0$\times$) & 1.70 (1.2$\times$) & \textbf{0.89 (2.3$\times$)} \\
    \midrule
    \multirow{2}{*}{Nrtv (10k)} & F1 $\uparrow$& 14.7 (100\%) & 2.0 (13.6\%) & \textbf{14.7 (100\%)} \\
    & Time (s) & 6.82 (1.0$\times$) & 4.55 (1.5$\times$) & \textbf{2.85 (2.4$\times$)} \\
    \midrule
    \multirow{2}{*}{QALT (7k)} & AC $\uparrow$& 48.8 (100\%) & 6.8 (13.9\%) & \textbf{48.6 (99.6\%)} \\
    & Time (s) & 3.76 (1.0$\times$) & 2.09 (1.8$\times$) & \textbf{1.26 (3.0$\times$)} \\
    \midrule
    \multirow{2}{*}{MuSQ (3k)} & F1 $\uparrow$& 18.6 (100\%) & 16.9 (90.9\%) & \textbf{18.5 (99.5\%)} \\
    & Time (s) & 0.70 (1.0$\times$) & 0.63 (1.1$\times$) & \textbf{0.37 (1.9$\times$)} \\
    \midrule
    \multirow{2}{*}{SpDg (7.5k)} & ES $\uparrow$ & 42.5 (100\%) & 2.9 (6.8\%) & \textbf{42.3 (99.5\%)} \\
    & Time (s) & 4.43 (1.0$\times$) & 2.22 (2.0$\times$) & \textbf{1.47 (3.0$\times$)} \\
    \midrule
    \multirow{2}{*}{BkSS (7.5k)} & $C_{idx} \uparrow$ & 19.5 (100\%) & 11.7 (60.0\%) & \textbf{19.3 (99.0\%)} \\
    & Time (s) & 4.52 (1.0$\times$) & 2.26 (2.0$\times$) & \textbf{1.55 (2.9$\times$)} \\
    \midrule
    \midrule
    \multirow{2}{*}{Avg. (7.5k)} & / $\uparrow$ & 23.9 (100\%) & 7.8 (32.5\%) & \textbf{23.8 (99.6\%)} \\
    & Time (s) & 4.38 (1.0$\times$) & 2.92 (1.5$\times$) & \textbf{1.60 (2.7$\times$)} \\
    \bottomrule
  \end{tabular}
  }
  \label{tab:zeroscroll}
\end{table}

\begin{figure}
  \centering
\includegraphics[width=0.99\linewidth]{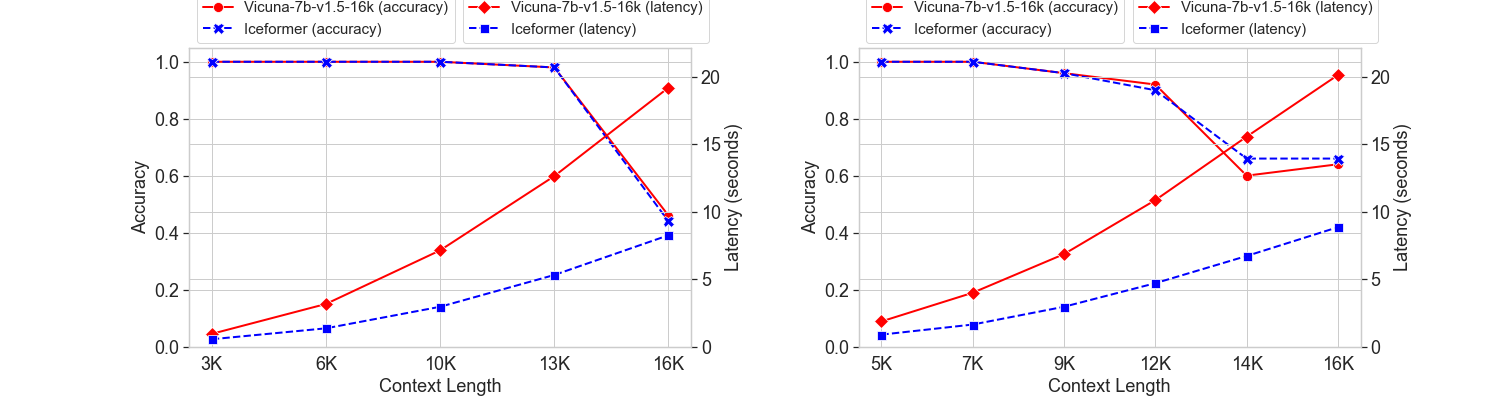}
  \caption{Scalability analysis for \toolName on the LongEval benchmark. The left figure shows the results of the topic retrieval task; the right figure shows the results of the line retrieval task. X-axis: length of the input prompt; Y-axis (Left): retrieval accuracy; Y-axis (Right): averaged process wall clock time (second) of the attention module.}
  \label{fig:LongEval}
\end{figure}

\paragraph{LongEval Results \& Scalability Analysis.}
To provide a more comprehensive analysis of \toolName's scalability in the LLM setting, we conducted additional experiments on the LongEval benchmark~\citep{longchat2023}, which is designed to measure long-context performance and consists of two tasks: topic retrieval task with prompt length varying from 3k to 16k, and line retrieval task with prompt length varying from 5k to 16k. In Figure~\ref{fig:LongEval}, we present the averaged latency of the attention module corresponding to different input prompt length as well as the inference accuracy using the \revise{vanilla Vicuna-7b-v1.5-16k model} and \toolName. From the figure, \toolName can achieve nearly identical inference accuracy compared with the vanilla Vicuna-7b-v1.5-16k. Notably, as the prompt length increases, there is a corresponding increase in the inference latency for both methods and for both tasks. However, even with very long prompt lengths, \toolName maintains its scalability and consistently outperforms the vanilla Transformer. Furthermore, as the length of the prompt increases, the difference in the latency between \toolName and the vanilla Transformer becomes larger, demonstrating the superior scalability and efficiency of \toolName in the context of LLMs. 
% Notably, \toolName can sometimes achieve even better accuracy in the line retrieval
% task. We treat these results as normal since Prioritized DCI

\section{Conclusion}
In this paper, we present \toolName, a new method for improving the inference time efficiency of pretrained Transformers on the CPU. Notably, in contrast to other methods, \toolName does not require retraining, does not require special constraints imposed on the attention mechanism and simultaneously achieves high accuracy and fast inference. These advantages make \toolName very well-suited to LLM deployment on CPUs, especially when the LLM needs to handle very long sequences as input. The experimental findings on three benchmarks compellingly illustrate the effectiveness of our approach in reducing the quadratic time and space complexity of Transformers both in cases with bi-directional and causal attention mechanisms.

% \subsubsection*{Author Contributions}
% If you'd like to, you may include  a section for author contributions as is done
% in many journals. This is optional and at the discretion of the authors.

% \subsubsection*{Acknowledgments}
% Use unnumbered third level headings for the acknowledgments. All
% acknowledgments, including those to funding agencies, go at the end of the paper.

\bibliography{iclr2024_conference}

\begin{thebibliography}{32}
\providecommand{\natexlab}[1]{#1}
\providecommand{\url}[1]{\texttt{#1}}
\expandafter\ifx\csname urlstyle\endcsname\relax
  \providecommand{\doi}[1]{doi: #1}\else
  \providecommand{\doi}{doi: \begingroup \urlstyle{rm}\Url}\fi

\bibitem[Ainslie et~al.(2020)Ainslie, Ontanon, Alberti, Cvicek, Fisher, Pham,
  Ravula, Sanghai, Wang, and Yang]{ainslie2020etc}
Joshua Ainslie, Santiago Ontanon, Chris Alberti, Vaclav Cvicek, Zachary Fisher,
  Philip Pham, Anirudh Ravula, Sumit Sanghai, Qifan Wang, and Li~Yang.
\newblock Etc: Encoding long and structured inputs in transformers.
\newblock \emph{arXiv preprint arXiv:2004.08483}, 2020.

\bibitem[Andoni et~al.(2015)Andoni, Indyk, Laarhoven, Razenshteyn, and
  Schmidt]{andoni2015practical}
Alexandr Andoni, Piotr Indyk, Thijs Laarhoven, Ilya Razenshteyn, and Ludwig
  Schmidt.
\newblock Practical and optimal lsh for angular distance.
\newblock \emph{Advances in neural information processing systems}, 28, 2015.

\bibitem[Anthropic(2023)]{anthropic-100k-context-windows}
Anthropic.
\newblock 100k context windows, 2023.
\newblock URL \url{https://www.anthropic.com/index/100k-context-windows}.

\bibitem[Aum{\"u}ller et~al.(2017)Aum{\"u}ller, Bernhardsson, and
  Faithfull]{aumuller2017ann}
Martin Aum{\"u}ller, Erik Bernhardsson, and Alexander Faithfull.
\newblock Ann-benchmarks: A benchmarking tool for approximate nearest neighbor
  algorithms.
\newblock In \emph{International conference on similarity search and
  applications}, pp.\  34--49. Springer, 2017.

\bibitem[Beltagy et~al.(2020)Beltagy, Peters, and Cohan]{beltagy2020longformer}
Iz~Beltagy, Matthew~E Peters, and Arman Cohan.
\newblock Longformer: The long-document transformer.
\newblock \emph{arXiv preprint arXiv:2004.05150}, 2020.

\bibitem[Child et~al.(2019)Child, Gray, Radford, and
  Sutskever]{child2019generating}
Rewon Child, Scott Gray, Alec Radford, and Ilya Sutskever.
\newblock Generating long sequences with sparse transformers.
\newblock \emph{arXiv preprint arXiv:1904.10509}, 2019.

\bibitem[Choromanski et~al.(2020)Choromanski, Likhosherstov, Dohan, Song, Gane,
  Sarlos, Hawkins, Davis, Mohiuddin, Kaiser, et~al.]{choromanski2020rethinking}
Krzysztof Choromanski, Valerii Likhosherstov, David Dohan, Xingyou Song,
  Andreea Gane, Tamas Sarlos, Peter Hawkins, Jared Davis, Afroz Mohiuddin,
  Lukasz Kaiser, et~al.
\newblock Rethinking attention with performers.
\newblock \emph{arXiv preprint arXiv:2009.14794}, 2020.

\bibitem[Dao et~al.(2022)Dao, Fu, Ermon, Rudra, and
  R{\'e}]{dao2022flashattention}
Tri Dao, Dan Fu, Stefano Ermon, Atri Rudra, and Christopher R{\'e}.
\newblock Flashattention: Fast and memory-efficient exact attention with
  io-awareness.
\newblock \emph{Advances in Neural Information Processing Systems},
  35:\penalty0 16344--16359, 2022.

\bibitem[Dice \& Kogan(2021)Dice and Kogan]{dice2021optimizing}
Dave Dice and Alex Kogan.
\newblock Optimizing inference performance of transformers on cpus.
\newblock \emph{arXiv preprint arXiv:2102.06621}, 2021.

\bibitem[Gupta et~al.(2021)Gupta, Dar, Goodman, Ciprut, and
  Berant]{gupta2021memory}
Ankit Gupta, Guy Dar, Shaya Goodman, David Ciprut, and Jonathan Berant.
\newblock Memory-efficient transformers via top-$ k $ attention.
\newblock \emph{arXiv preprint arXiv:2106.06899}, 2021.

\bibitem[Indyk \& Motwani(1998)Indyk and Motwani]{indyk1998approximate}
Piotr Indyk and Rajeev Motwani.
\newblock Approximate nearest neighbors: towards removing the curse of
  dimensionality.
\newblock In \emph{Proceedings of the thirtieth annual ACM symposium on Theory
  of computing}, pp.\  604--613, 1998.

\bibitem[Krizhevsky et~al.(2009)Krizhevsky, Hinton,
  et~al.]{krizhevsky2009learning}
Alex Krizhevsky, Geoffrey Hinton, et~al.
\newblock Learning multiple layers of features from tiny images.
\newblock 2009.

\bibitem[Li et~al.(2023)Li, Shao, Xie, Sheng, Zheng, Gonzalez, Stoica, Ma, and
  Zhang]{longchat2023}
Dacheng Li, Rulin Shao, Anze Xie, Ying Sheng, Lianmin Zheng, Joseph Gonzalez,
  Ion Stoica, Xuezhe Ma, and Hao Zhang.
\newblock How long can context length of open-source {LLM}s truly promise?
\newblock In \emph{NeurIPS 2023 Workshop on Instruction Tuning and Instruction
  Following}, 2023.
\newblock URL \url{https://openreview.net/forum?id=LywifFNXV5}.

\bibitem[Li \& Malik(2017)Li and Malik]{li2017fast}
Ke~Li and Jitendra Malik.
\newblock Fast k-nearest neighbour search via prioritized dci.
\newblock In \emph{International conference on machine learning}, pp.\
  2081--2090. PMLR, 2017.

\bibitem[Linsley et~al.(2018)Linsley, Kim, Veerabadran, Windolf, and
  Serre]{linsley2018learning}
Drew Linsley, Junkyung Kim, Vijay Veerabadran, Charles Windolf, and Thomas
  Serre.
\newblock Learning long-range spatial dependencies with horizontal gated
  recurrent units.
\newblock \emph{Advances in neural information processing systems}, 31, 2018.

\bibitem[Maas et~al.(2011)Maas, Daly, Pham, Huang, Ng, and
  Potts]{maas-EtAl:2011:ACL-HLT2011}
Andrew~L. Maas, Raymond~E. Daly, Peter~T. Pham, Dan Huang, Andrew~Y. Ng, and
  Christopher Potts.
\newblock Learning word vectors for sentiment analysis.
\newblock In \emph{Proceedings of the 49th Annual Meeting of the Association
  for Computational Linguistics: Human Language Technologies}, pp.\  142--150,
  Portland, Oregon, USA, June 2011. Association for Computational Linguistics.
\newblock URL \url{http://www.aclweb.org/anthology/P11-1015}.

\bibitem[Nangia \& Bowman(2018)Nangia and Bowman]{nangia2018listops}
Nikita Nangia and Samuel~R Bowman.
\newblock Listops: A diagnostic dataset for latent tree learning.
\newblock \emph{arXiv preprint arXiv:1804.06028}, 2018.

\bibitem[Neyshabur \& Srebro(2015)Neyshabur and Srebro]{neyshabur2015symmetric}
Behnam Neyshabur and Nathan Srebro.
\newblock On symmetric and asymmetric lshs for inner product search.
\newblock In \emph{International Conference on Machine Learning}, pp.\
  1926--1934. PMLR, 2015.

\bibitem[Nikita et~al.(2020)Nikita, Lukasz, Anselm, et~al.]{nikita2020reformer}
Kitaev Nikita, Kaiser Lukasz, Levskaya Anselm, et~al.
\newblock Reformer: The efficient transformer.
\newblock In \emph{Proceedings of International Conference on Learning
  Representations (ICLR)}, 2020.

\bibitem[OpenAI(2023)]{openai-gpt-4}
OpenAI.
\newblock Openai gpt-4, 2023.
\newblock URL \url{https://openai.com/gpt-4}.

\bibitem[Peng et~al.(2021)Peng, Pappas, Yogatama, Schwartz, Smith, and
  Kong]{peng2021random}
Hao Peng, Nikolaos Pappas, Dani Yogatama, Roy Schwartz, Noah~A Smith, and
  Lingpeng Kong.
\newblock Random feature attention.
\newblock \emph{arXiv preprint arXiv:2103.02143}, 2021.

\bibitem[Radev et~al.(2013)Radev, Muthukrishnan, Qazvinian, and
  Abu-Jbara]{radev2013acl}
Dragomir~R Radev, Pradeep Muthukrishnan, Vahed Qazvinian, and Amjad Abu-Jbara.
\newblock The acl anthology network corpus.
\newblock \emph{Language Resources and Evaluation}, 47:\penalty0 919--944,
  2013.

\bibitem[Shaham et~al.(2023)Shaham, Ivgi, Efrat, Berant, and
  Levy]{shaham2023zeroscrolls}
Uri Shaham, Maor Ivgi, Avia Efrat, Jonathan Berant, and Omer Levy.
\newblock Zeroscrolls: A zero-shot benchmark for long text understanding, 2023.

\bibitem[Tay et~al.(2020)Tay, Dehghani, Abnar, Shen, Bahri, Pham, Rao, Yang,
  Ruder, and Metzler]{tay2020long}
Yi~Tay, Mostafa Dehghani, Samira Abnar, Yikang Shen, Dara Bahri, Philip Pham,
  Jinfeng Rao, Liu Yang, Sebastian Ruder, and Donald Metzler.
\newblock Long range arena: A benchmark for efficient transformers.
\newblock \emph{arXiv preprint arXiv:2011.04006}, 2020.

\bibitem[Touvron et~al.(2023)Touvron, Martin, Stone, Albert, Almahairi, Babaei,
  Bashlykov, Batra, Bhargava, Bhosale, et~al.]{touvron2023llama2}
Hugo Touvron, Louis Martin, Kevin Stone, Peter Albert, Amjad Almahairi, Yasmine
  Babaei, Nikolay Bashlykov, Soumya Batra, Prajjwal Bhargava, Shruti Bhosale,
  et~al.
\newblock Llama 2: Open foundation and fine-tuned chat models.
\newblock \emph{arXiv preprint arXiv:2307.09288}, 2023.

\bibitem[Vaswani et~al.(2017)Vaswani, Shazeer, Parmar, Uszkoreit, Jones, Gomez,
  Kaiser, and Polosukhin]{vaswani2017attention}
Ashish Vaswani, Noam Shazeer, Niki Parmar, Jakob Uszkoreit, Llion Jones,
  Aidan~N Gomez, {\L}ukasz Kaiser, and Illia Polosukhin.
\newblock Attention is all you need.
\newblock In \emph{Advances in neural information processing systems}, pp.\
  5998--6008, 2017.

\bibitem[Wang et~al.(2020)Wang, Li, Khabsa, Fang, and Ma]{wang2020linformer}
Sinong Wang, Belinda~Z Li, Madian Khabsa, Han Fang, and Hao Ma.
\newblock Linformer: Self-attention with linear complexity.
\newblock \emph{arXiv preprint arXiv:2006.04768}, 2020.

\bibitem[Xiong et~al.(2021)Xiong, Zeng, Chakraborty, Tan, Fung, Li, and
  Singh]{xiong2021nystromformer}
Yunyang Xiong, Zhanpeng Zeng, Rudrasis Chakraborty, Mingxing Tan, Glenn Fung,
  Yin Li, and Vikas Singh.
\newblock Nystr{\"o}mformer: A nystr{\"o}m-based algorithm for approximating
  self-attention.
\newblock In \emph{Proceedings of the AAAI Conference on Artificial
  Intelligence}, volume~35, pp.\  14138--14148, 2021.

\bibitem[Zhang et~al.(2023)Zhang, Sheng, Zhou, Chen, Zheng, Cai, Song, Tian,
  R{\'e}, Barrett, et~al.]{zhang2023h}
Zhenyu Zhang, Ying Sheng, Tianyi Zhou, Tianlong Chen, Lianmin Zheng, Ruisi Cai,
  Zhao Song, Yuandong Tian, Christopher R{\'e}, Clark Barrett, et~al.
\newblock H $ \_2 $ o: Heavy-hitter oracle for efficient generative inference
  of large language models.
\newblock \emph{arXiv preprint arXiv:2306.14048}, 2023.

\bibitem[Zheng et~al.(2023)Zheng, Chiang, Sheng, Zhuang, Wu, Zhuang, Lin, Li,
  Li, Xing, Zhang, Gonzalez, and Stoica]{zheng2023judging}
Lianmin Zheng, Wei-Lin Chiang, Ying Sheng, Siyuan Zhuang, Zhanghao Wu, Yonghao
  Zhuang, Zi~Lin, Zhuohan Li, Dacheng Li, Eric.~P Xing, Hao Zhang, Joseph~E.
  Gonzalez, and Ion Stoica.
\newblock Judging llm-as-a-judge with mt-bench and chatbot arena, 2023.

\bibitem[Zheng et~al.(2022)Zheng, Wang, and Kong]{zheng2022linear}
Lin Zheng, Chong Wang, and Lingpeng Kong.
\newblock Linear complexity randomized self-attention mechanism.
\newblock In \emph{International Conference on Machine Learning}, pp.\
  27011--27041. PMLR, 2022.

\bibitem[Zhu \& Soricut(2021)Zhu and Soricut]{zhu-soricut-2021-h}
Zhenhai Zhu and Radu Soricut.
\newblock {H}-transformer-1{D}: Fast one-dimensional hierarchical attention for
  sequences.
\newblock In \emph{Proceedings of the 59th Annual Meeting of the Association
  for Computational Linguistics and the 11th International Joint Conference on
  Natural Language Processing (Volume 1: Long Papers)}, pp.\  3801--3815,
  Online, August 2021. Association for Computational Linguistics.
\newblock \doi{10.18653/v1/2021.acl-long.294}.
\newblock URL \url{https://aclanthology.org/2021.acl-long.294}.

\end{thebibliography}
\bibliographystyle{iclr2024_conference}

\newpage
\appendix
\section{\revise{Pseudocode for \toolName}}\label{sec:code}
\revise{We provide the pseudocode below for \toolName. For completeness, we also include the pseudocode of Prioritized DCI, which we adapted from \citep{li2017fast} and added our modifications. In the following pseudocode below, $T_K$ and $T_Q$ are the key and query embedding functions, respectively. In our implementation, we implement a recursive version of the algorithm, which offers faster performance in practice.} 

%**We will release the code upon acceptance.
% Code will be released upon acceptance.

\begin{algorithm}
\footnotesize
\caption{\toolName attention}
\label{alg_atten}
\begin{algorithmic}
\Require A set $S_q$ of $n$ points $q^{1},\ldots,q^{n} \in \mathbb{R}^{d}$, a set $S_k$ of $n$ keys $k^{1},\ldots,k^{n} \in \mathbb{R}^{d}$, a set $S_v$ of $n$ values $v^{1},\ldots,v^{n} \in \mathbb{R}^{d'}$, the number of simple indices $m$ that constitute a composite index, the number of composite indices $L$, the number of points to retrieve $k_{0}$ and the number of points to visit $k_{1}$ in each composite index
\Function{Iceformer\_Attention}{$S_q,S_k,S_v,m,L,k_{0},k_{1}$}
    \State $\mathbf{O} \gets $ zero matrix $\in \mathbb{R}^{n \times d'} $ with rows $\mathbf{o}_i \in \mathbb{R}^{d'} $
    \State $S_{p} \gets $ CONSTRUCT$(S_k,m,L)$
    \For{$i = 1$ \textbf{to} $n$}
        \State $S_{l} \gets $ QUERY$(q^{i},S_p,k_{0},k_{1})$
        \For{$j = 1$ \textbf{to} $n$}
            \If{$j \in S_{l}$}
                \State $\tilde{s}_{i j} \gets \frac{{q}_i^\top {k}_j}{\sqrt{d}}$
            \Else
                \State $\tilde{s}_{i j} \gets 0$
            \EndIf
        \EndFor
        \For{$j = 1$ \textbf{to} $n$}
            \If{$j \in S_{l}$}
                \State $\tilde{a}_{i,j} \gets \operatorname{ softmax}_j\left(\tilde{s}_{i j}\right)=\frac{\exp \left(\tilde{s}_{i,j}\right)}{\sum_{k \in S_{l}} \exp \left(\tilde{s}_{i,k}\right)}$
            \Else
                \State $\tilde{a}_{i j} \gets 0$
            \EndIf
        \EndFor
        \State
        $\mathbf{o}_i  \gets \sum_{j\in S_{l}} \tilde{a}_{i,j} {v}_j$
        
    \EndFor
    \State \Return $\mathbf{O}$
\EndFunction
\end{algorithmic}
\normalsize
\end{algorithm}

\begin{algorithm}
\footnotesize
\caption{Data structure construction procedure}
\label{alg_construct}
\begin{algorithmic}
\Require A dataset $D$ of $n$ points $k^{1},\ldots,k^{n}$, the number of simple indices $m$ that constitute a composite index and the number of composite indices $L$
\Function{Construct}{$D,m,L$}
    \State $\{u_{jl}\}_{j \in [m], l \in [L]} \gets mL$ random unit vectors in $\mathbb{R}^{d}$
    \State $\{T_{jl}\}_{j \in [m], l \in [L]} \gets mL$ empty binary search trees or skip lists
    % \State 
    \For{$j = 1$ \textbf{to} $m$}
        \For{$l = 1$ \textbf{to} $L$}
            \For{$i = 1$ \textbf{to} $n$}
                \State $\overline{k}^{i}_{jl} \gets \langle T_K(k^{i}),u_{jl}\rangle$
                \State Insert $(\overline{k}^{i}_{jl},i)$ into $T_{jl}$ with $\overline{k}^{i}_{jl}$ being the key and $i$ being the value
            \EndFor
        \EndFor
    \EndFor
    \State \Return $\{(T_{jl},u_{jl})\}_{j \in [m], l \in [L]}$
\EndFunction
\end{algorithmic}
\normalsize
\end{algorithm}

\begin{algorithm}[H]
\footnotesize
\caption{$k$-nearest neighbour querying procedure}
\label{alg_query}
\begin{algorithmic}
\Require Query point $q$ in $\mathbb{R}^{d}$, binary search trees/skip lists and their associated projection vectors $\{(T_{jl},u_{jl})\}_{j \in [m], l \in [L]}$, the number of points to retrieve $k_{0}$ and the number of points to visit $k_{1}$ in each composite index
\Function{Query}{$q,\{(T_{jl},u_{jl})\}_{j,l},k_{0},k_{1}$}
    \State $C_{l} \gets$ array of size $n$ with entries initialized to 0\; $\forall l \in [L]$
    \State $\overline{q}_{jl} \gets \langle T_Q(q),u_{jl}\rangle\; \forall j \in [m], l \in [L]$
    \State $S_{l} \gets \emptyset \; \forall l \in [L]$
    \State $P_{l} \gets$ empty priority queue $\; \forall l \in [L]$
    \For{$l = 1$ \textbf{to} $L$}
            \For{$j = 1$ \textbf{to} $m$}
                \State $(\overline{p}_{jl}^{(1)},h_{jl}^{(1)}) \gets $ the node in $T_{jl}$ whose key is the closest to $\overline{q}_{jl}$
                \State Insert $(\overline{p}_{jl}^{(1)},h_{jl}^{(1)}) $ with priority $-|\overline{p}_{jl}^{(1)} - \overline{q}_{jl}|$ into $P_{l}$
            \EndFor
    \EndFor
        \For{$i' = 1$ \textbf{to} $k_{1} - 1$}
            \For{$l = 1$ \textbf{to} $L$}
                \If{$\left|S_{l}\right| < k_{0}$}
                    \State $(\overline{p}_{jl}^{(i)},h_{jl}^{(i)}) \gets $ the node with the highest priority in $P_{l}$
                    \State Remove $(\overline{p}_{jl}^{(i)},h_{jl}^{(i)})$ from $P_{l}$ and insert the node in $T_{jl}$ whose key is the next closest to $\overline{q}_{jl}$, 
                    \State \quad which is denoted as $(\overline{p}_{jl}^{(i+1)},h_{jl}^{(i+1)})$, with priority $-|\overline{p}_{jl}^{(i+1)} - \overline{q}_{jl}|$ into $P_{l}$
                    \State $C_{l}[h_{jl}^{(i)}] \gets C_{l}[h_{jl}^{(i)}] + 1$
                    \If{$C_{l}[h_{jl}^{(i)}] = m$}
                        \State $S_{l} \gets S_{l} \cup \{h_{jl}^{(i)}\}$
                    \EndIf
                \EndIf
            \EndFor
        \EndFor
    \State \Return $k$ points in $\bigcup_{l\in[L]}S_{l}$ that have the maximum inner-product value with $q$
\EndFunction
\end{algorithmic}
\normalsize
\end{algorithm}

\section{Proofs}
\subsection{Proof 1}\label{proof1}
Here, we provide the full step-by-step derivation of the mathematical equivalence between conducting $k$-nearest neighbour search on normalized keys and identifying the keys that obtain the highest attention weight.
\begin{align}
\arg\max_j a_{i,j} & = \arg\max_j \mathrm{softmax}_j\left( \left\{\frac{\mathbf{q}_i^\top \mathbf{k}_{j'}}{\sqrt{d}}\right\}_{j'=1}^{m} \right) \\
& = \arg\max_j \frac{\mathbf{q}_i^\top \mathbf{k}_{j}}{\sqrt{d}} \\
%& = \arg\max_j \mathbf{q}_i^\top \mathbf{k}_{j} \\
& = \arg\min_j \Vert\mathbf{q}_i\Vert_2^2 -  2\mathbf{q}_i^\top\mathbf{k}_{j} + 1
\end{align}

Since $\Vert \mathbf{k}_{j'} \Vert_2 = 1$ for all $j'$, $\Vert\mathbf{q}_i\Vert_2^2 -  2\mathbf{q}_i^\top\mathbf{k}_{j} + 1 = \Vert\mathbf{q}_i\Vert_2^2 -  2\mathbf{q}_i^\top\mathbf{k}_{j} + \Vert\mathbf{k}_{j}\Vert_2^2 = \Vert \mathbf{q}_i - \mathbf{k}_{j} \Vert_2^2$, 
\begin{align}
\arg\max_j a_{i,j} & = \arg\min_j \Vert\mathbf{q}_i\Vert_2^2 -  2\mathbf{q}_i^\top\mathbf{k}_{j} + 1 \\
& = \arg\min_j \Vert \mathbf{q}_i - \mathbf{k}_{j} \Vert_2^2
\end{align}

\subsection{Proof 2}\label{proof2}
Here, we provide the full step-by-step derivation of the result in~\ref{sec:math} establishing the mathematical equivalence between conducting $k$-nearest neighbour search on transformed keys and identifying the keys that obtain the highest attention weight.
\begin{align}
\arg\max_j a_{i,j} & = \arg\max_j \mathrm{softmax}_j\left( \left\{\frac{\mathbf{q}_i^\top \mathbf{k}_{j'}}{\sqrt{d}}\right\}_{j'=1}^{m} \right) \\
& = \arg\max_j \frac{\mathbf{q}_i^\top \mathbf{k}_{j}}{\sqrt{d}} \\
& = \arg\max_j \mathbf{q}_i^\top \mathbf{k}_{j} \\
& = \arg\min_j -  2\mathbf{q}_i^\top\mathbf{k}_{j} \\
& = \arg\min_j 2 -  2\mathbf{q}_i^\top\mathbf{k}_{j} / c\Vert \mathbf{q}_i \Vert_2 \label{eq:dist_transformed}\\
& = \arg\min_j 1 -  2\mathbf{q}_i^\top\mathbf{k}_{j} / c\Vert \mathbf{q}_i \Vert_2 + 1 \\
& = \arg\min_j \Vert \mathbf{q}_i \Vert_2^2 / \Vert \mathbf{q}_i \Vert_2^2 -  2\mathbf{q}_i^\top\mathbf{k}_{j} / c\Vert \mathbf{q}_i \Vert_2 + \Vert \mathbf{k}_{j} \Vert_2^2 / c^2 + 1 - \|\mathbf{k}_{j}\|_2^2 / c^2 \\
& = \arg\min_j \mathbf{q}_i^\top \mathbf{q}_i / \Vert \mathbf{q}_i \Vert_2^2 -  2\mathbf{q}_i^\top\mathbf{k}_{j} / c\Vert \mathbf{q}_i \Vert_2 + \mathbf{k}_{j}^\top \mathbf{k}_{j} / c^2 + 1 - \|\mathbf{k}_{j}\|_2^2 / c^2 \\
& = \arg\min_j (\mathbf{q}_i / \Vert \mathbf{q}_i \Vert_2  - \mathbf{k}_{j} / c)^\top (\mathbf{q}_i / \Vert \mathbf{q}_i \Vert_2 - \mathbf{k}_{j} / c) + 1 - \|\mathbf{k}_{j}\|_2^2 / c^2 \\
& = \arg\min_j \Vert \mathbf{q}_i / \Vert \mathbf{q}_i\Vert_2 - \mathbf{k}_{j} / c \Vert_2^2 + 1 - \|\mathbf{k}_{j}\|_2^2 / c^2 \\
& = \arg\min_j \Vert \mathbf{q}_i / \Vert \mathbf{q}_i\Vert_2 - \mathbf{k}_{j} / c \Vert_2^2 + \left(0 - \sqrt{1-\|\mathbf{k}_{j}\|_2^2 / c^2} \right)^2 \\
& = \arg\min_j \Vert T_Q(\mathbf{q}_i) - T_K(\mathbf{k}_{j}) \Vert_2^2 \\
& = \arg\min_j \Vert T_Q(\mathbf{q}_i) - T_K(\mathbf{k}_{j}) \Vert_2
\end{align}

\section{More Details on the LRA Experimental Setting}\label{sec:lra}
\subsection{Dataset Details}\label{sec:lra_data}
In our LRA experiments, for Retrieval, Text and Pathfinder ($64 \times 64$ version), we directly use the dataset from LRA codebase\footnote{https://github.com/google-research/long-range-arena/tree/main}. Because the original datasets for ListOps and Image only contain short sequences, we generate longer samples for ListOps using the same code from the LRA codebase with 4000 as the maximum length; for Image task, we use a version of the CIFAR-10 dataset super-resolved to $64 \times 64$\footnote{https://www.kaggle.com/datasets/joaopauloschuler/cifar10-64x64-resized-via-cai-super-resolution} instead of the original low-resolution $32 \times 32$ CIFAR-10 dataset. We follow the exact same train/test split as the original LRA paper~\citep{tay2020long}. The details of the LRA dataset is listed in Table~\ref{tab:lra}.
\begin{table}[htp]
\caption{LRA Dataset Details.}
\centering
\scalebox{0.63} {
\setlength{\tabcolsep}{4pt}
\begin{tabular}{c|c|c|c|c|c}
\toprule {Task}  &  {ListOps}  &  {Text} & {Retrieval} & {Image} & {Pathfinder}  \\
\midrule
\midrule
{Max length} & 3,991 & 4,000 & 4,000 & 4,096 & 4,096 \\
{Avg. length} & 2,232 &  1,267 & 3,917 & 4,096 & 4,096  \\
{number of classes} & 10 &  2 & 2 & 10 & 2  \\
{Accuracy by chance} &   0.100 & 0.500 & 0.500  & 0.100 & 0.500  \\
\bottomrule
\end{tabular}
}
\label{tab:lra}
\end{table}

\subsection{Base Model Configuration}\label{sec:setting}
We follow the experimental setup of prior work~\citep{zhu-soricut-2021-h} for training the base model. However, since we were not able to successfully train base Transformer models to satisfactory accuracy on Image and Pathfinder datasets using the original setting, we decreased the number of heads and layers for these two tasks. The details of the base model for each task are outlined in Table~\ref{tab:lra-model}.
\begin{table}[htp]
\caption{Configurations of the base models for different tasks.}
\centering
\scalebox{0.63} {
\setlength{\tabcolsep}{4pt}
\begin{tabular}{c|c|c|c|c|c}
\toprule {Task}  &  {ListOps}  &  {Text} & {Retrieval} & {Image} & {Pathfinder}  \\
\midrule
\midrule
{head embedding size} & 512 & 512 & 512 & 512 & 512  \\
{feed-forward size} & 2048 & 2048 & 2048 & 1024 & 1024  \\
{number of heads} &  8 & 8 & 8 & 4 & 4   \\
{number of layers} & 6 & 6 & 6 & 4 & 4   \\
\bottomrule
\end{tabular}
}
\label{tab:lra-model}
\end{table}

\subsection{Hyper-parameters for the Baselines and the Proposed Method}\label{sec:hyper}
For LARA and Nystr\"{o}mformer, we tuned the parameter \textit{num\_landmarks} by optimizing over the range \{64, 128, 256, 512, 1024\}. For H-Transformer-1D, we tuned the parameter \textit{block\_size} by optimizing over the range \{64, 128, 256, 512, 1024\}. For Reformer, we tuned the parameters \textit{num\_hash} and \textit{bucket\_size}: we considered the values of \textit{num\_hash} in range \{1, 2, 4\} and the values of \textit{bucket\_size} in range \{64, 128, 256, 512, 1024\}.For Longformer, Performer, and Linformer which require retraining, because of their poor performance, we choose hyper-parameter values that result in the least amount of approximation. For \toolName, we tuned the parameter \textit{top\_k} over the range \{3, 5, 8, 10, 15, 20\}. In general, a larger value for \textit{bucket\_size}, \textit{num\_landmarks}, \textit{block\_size}, or \textit{top\_k} indicates less aggressive approximation, meaning that the model performance is closer to that of the vanilla Transformer. We select the values of the hyper-parameters that lead to the best accuracy-time trade-off for each model, and list them in Table~\ref{tab:lra-hype}.

\begin{table}[htb]
\caption{Hyper-parameter settings for different methods.}
\centering
\scalebox{0.63} {
\setlength{\tabcolsep}{4pt}
\begin{tabular}{c|c|c|c|c|c|c}
\toprule {Method}  & {hyper-parameter}   &  {ListOps}  &  {Text} & {Retrieval} & {Image} &  {Pathfinder}  \\
\midrule
\midrule
\multirow{2}{*}{Reformer~\citep{nikita2020reformer}} & num\_hash  & 1  & 1 & 1  & 1 & / \\
& bucket\_size & 512 & 128 & 256 & 1024 & /  \\
\midrule
{LARA~\citep{zheng2022linear}} & num\_landmarks  &  256 & 256 & 512 & 1024 & 1024 \\
\midrule
{Nystr\"{o}mformer~\citep{xiong2021nystromformer}} &  num\_landmarks & 256 & 256 & 512 & 512 & 1024\\
\midrule
{H-Transformer-1D~\citep{zhu-soricut-2021-h}} & block\_size & 1024 & 512 & 1024 & 256 & 1024\\
\midrule
\revise{{Longformer~\citep{beltagy2020longformer}}} & \revise{attention\_window} & \revise{2048} & \revise{2048} & \revise{2048} & \revise{2048} & \revise{2048} \\
\midrule
\revise{{Performer~\citep{choromanski2020rethinking}}} & \revise{num\_rand\_features} & \revise{2048} & \revise{2048} & \revise{2048} & \revise{2048} & \revise{2048} \\
\midrule
\revise{{Linformer~\citep{wang2020linformer}}} & \revise{num\_proj\_dim} & \revise{2048} & \revise{2048} & \revise{2048} & \revise{2048} & \revise{2048} \\
\midrule
\midrule
% \multirow{2}{*}{\toolName}  
{\toolName} 
% & num\_layer & 4 & 3 & 4 & 4 & 2 \\
& top\_k & 8 & 3 & 10 & 10 & 10 \\
\midrule
% \multirow{2}{*}{\toolName (shared-QK)}  
{\toolName (shared-QK)} 
% & num\_layer & 4 & 3 & 4 & 4 & / \\
& top\_k & 10 & 3 & 10 & 20 & / \\
\bottomrule
\end{tabular}
}
\label{tab:lra-hype}
\end{table}

\section{Additional Experiments on LRA}
\paragraph{Approximation Quality.}
In order to assess how well various efficient Transformers approximate the outputs of the vanilla modified attention module, we measure the approximation error by computing the L2-norm of the difference between their attention module outputs and those of the standard vanilla attention module ($\mathbf{o}_i$ in Equation~\ref{eq:value_comb}). The averaged approximation errors for different efficient Transformers, utilizing the same hyper-parameter settings of Table~\ref{tab:lra-hype}, are summarized in Table~\ref{tab:3}. As indicated in the table, \toolName consistently achieves the lowest approximation errors across all LRA tasks, providing further evidence of its approximation efficacy.

\begin{table}[htp]
\caption{Quality of the approximation on LRA benchmark. The approximation error of the attention module output is reported for each method across all the tasks.}
\centering
\scalebox{0.63} {
\setlength{\tabcolsep}{4pt}
\begin{tabular}{c|c|c|c|c|c|c}
\toprule {Method}  & {shared-KQ}   &  {ListOps}  &  {Text} & {Retrieval} & {Image} & {Pathfinder}  \\
\midrule
\midrule
{Reformer~\citep{nikita2020reformer}} & \cmark & 3.823 & 3.926 & 5.452 & 2.130 & /\\
{LARA~\citep{zheng2022linear}} & \xmark   & 2.395 & 9.456 & 10.025 & 22.066 & 9.261 \\
{Nystr\"{o}mformer~\citep{xiong2021nystromformer}} & \xmark & 5.758 & 10.269 & 6.523 & 18.789 & 10.442  \\
{H-Transformer-1D~\citep{zhu-soricut-2021-h}} & \xmark & 6.110 & 10.605 & 5.676  & 53.926 & 12.228  \\
\midrule
\midrule
\multirow{2}{*}{\toolName (ours)}  
& \xmark  & \textbf{2.140} & \textbf{3.891} & \textbf{1.825} & \textbf{6.873} & \textbf{8.749} \\
& \cmark & \textbf{1.562} & \textbf{1.686} & \textbf{2.499} & \textbf{2.127} & /\\
\bottomrule
\end{tabular}
}
\label{tab:3}
\end{table}

\paragraph{\revise{Visualization of Tables 1\&2.}}

\begin{figure}[H]
  \centering
\includegraphics[width=1\linewidth]{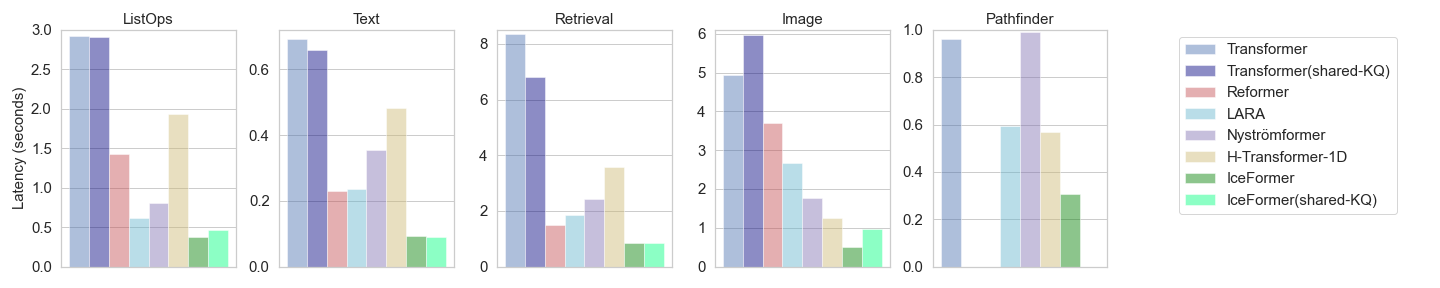}
  \caption{\revise{The inference latency of \toolName and the baselines (vanilla Transformer, Reformer, LARA, Nystr\"{o}mer, H-Transformer-1D) on the LRA benchmark (the smaller the better).}}
  \label{fig:viz_tab_1}
\end{figure}

\begin{figure}[H]
  \centering
\includegraphics[width=1\linewidth]{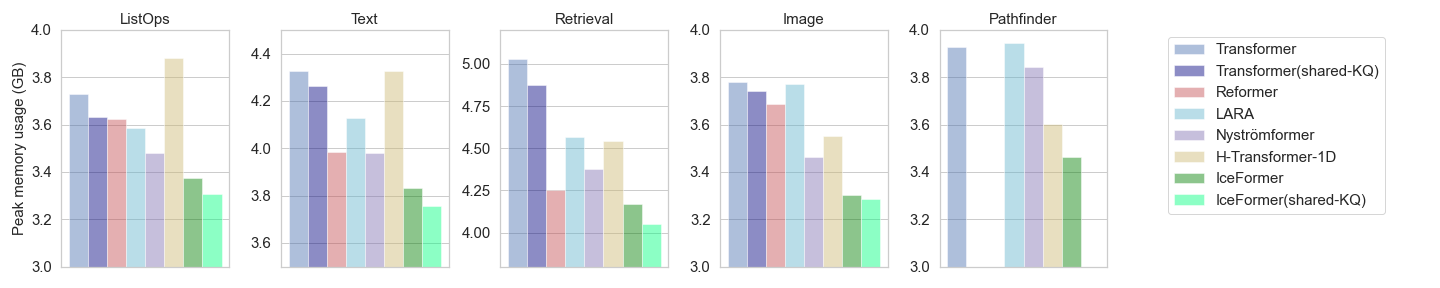}
  \caption{\revise{Peak memory usage (GB) on the LRA benchmark. The peak memory usage is the total memory usage of the whole program, which includes the memory for the Prioritized DCI database/index (the smaller the better).}}
  \label{fig:viz_tab_2}
\end{figure}

\revise{The results from Table 1 and Table 2 are visually represented in Figures~\ref{fig:viz_tab_1} and \ref{fig:viz_tab_2}, respectively.}

% \paragraph{Speed \& Accuracy Trade-off.}
% For efficient Transformers, increasing the extent of approximation generally improves model efficiency but can lead to a decrease in prediction performance. Here, we study how the extent of approximation affects inference speed and accuracy by varying the hyperparameters of all non-shared-KQ efficient Transformers. We select Retrieval and Text as examples and present the results in Figure~\ref{fig:3}. From the figure, we observe that the Time-Accuracy line of \toolName is the highest in both tasks, which indicates that \toolName can achieve accuracy close to that of the vanilla Transformer with the least time sacrifice across all the compared methods. 
% \begin{figure}[htp]
%   \begin{center}
%     \subfigure{\includegraphics[scale=0.33]{retrieval.png}}
%     \subfigure{\includegraphics[scale=0.33]{text.png}} \\
%   \end{center}
%   \caption{Changes of speed and accuracy on Retrieval (Left) and Text (Right) tasks. X-axis: the averaged wall clock time of attention module. Y-axis: the prediction accuracy of the model.}
%   \label{fig:3}
% \end{figure}

\section{More Details on the LLM Experiment}
\subsection{LLM Experiment Setting} \label{sec:llm}
We use vicuna-7b-v1.5-16k as the tested LLM in section~\ref{sec:llm-exp}. It contains 32 attention layers, each of which contains 32 attention heads with dimensionality equals to 128. Its maximum input sequence length is 16,384. We observe varying levels of sparsity across different layers of LLMs, and this sparsity remains consistent across different prompts. Therefore, in all the LLMs experiments in section~\ref{sec:llm-exp}, we apply \toolName to approximate relatively sparse layers ranging from the sixteenth to the thirty-first in vicuna-7b-v1.5-16k. This selection encompassed a total of 16 layers, equivalent to half of the total number of layers in the model.

The $k$ in the $k$-NNS of \toolName for each task of the ZeroSCROLLS benchmark and the LongEval benchmark is defined as: 
\begin{equation}
    k = \max(\min( \lfloor n * \alpha  \rfloor, 50), 30)
\end{equation}

where $n$ is the number of input tokens, $\lfloor x \rfloor$ is the floor function, and $\alpha$ is a hyper-parameter set by the users. In the ZeroSCROLLS benchmark, we set $\alpha$ equals to 4e-3 for tasks SSFD and QMsm; 5e-3 for tasks GvRp, SQAL, Qspr, Nrtv, MuSQ and BkSS; 6e-3 for tasks QALT and SpDg. In the LongEval benchmark, we set $\alpha$ equals to 5e-3 for all the settings of both two tasks.

% \begin{table}[h]
%   \caption{Details of the configuration of LLMs.}
%   \centering
%   \scalebox{0.72} {
%   \begin{tabular}{c|c|c|c}
%     \toprule
%     % \multicolumn{2}{c}{Part}                   \\
%     % \cmidrule(r){1-2}
%       & {StableLM-7B}  & {StarCoder} & {MPT-7B-StoryWriter} \\
%     \midrule
%     \midrule
%     {head embedding size} & 128 & 128 & 128 \\
%     {number of heads} & 48 & 48  & 32 \\
%     {number of layers} & 16  & 40 & 32 \\
%     {maximum sequence length} & 4096  & 8192 & 65536 \\
%     \bottomrule
%   \end{tabular}
%   }
%   \label{tab:llm-conf}
% \end{table}

% \subsection{Details on Applying \toolName to LLM}
% We observe varying levels of sparsity across different layers of LLMs, and this sparsity remains consistent across different prompts. Our analysis reveals that the majority of layers in the three LLMs we tried exhibit sparsity, with the exception of the first or last $X$ layers (where $X \leq 4$). In our experiments, we apply \toolName only to the layers where the sum of the top-20 probabilities exceeds 70\%. Specifically, we approximate layers ranging from the fifth to the sixteenth (the last layer) for StableLM-7B; layers ranging from the third to the fourtieth (the last layer) for StarCoder; layers ranging from the fourth to the thirty-first for MPT-7B-StoryWriter.

\subsection{Causal Masks and Other Inference-Time Efficient Transformers}\label{appendix:landmark}
In the main paper, we did not compare \toolName with LARA and Nystr\"{o}mformer on LLMs. In this section, we elaborate on the problems of causal masks for these two methods.

Most random-feature-based models such as LARA and Nystr\"{o}mformer group different tokens into different clusters, known as \emph{landmarks}. In order to enable causal masking in these models, not only does the masking need to be applied at the landmark level to prevent the leakage of information from future tokens, an additional set of masks is also required to mask out different numbers of tokens within the same landmark for different queries. The latter is not supported natively and is especially difficult to implement. As a result, it is difficult to apply LARA and Nystr\"{o}mformer to models that have causal masks.

\section{Text outputs of \toolName + LLM}
\begin{figure}[htb]
  \centering
\includegraphics[width=1.0\linewidth]{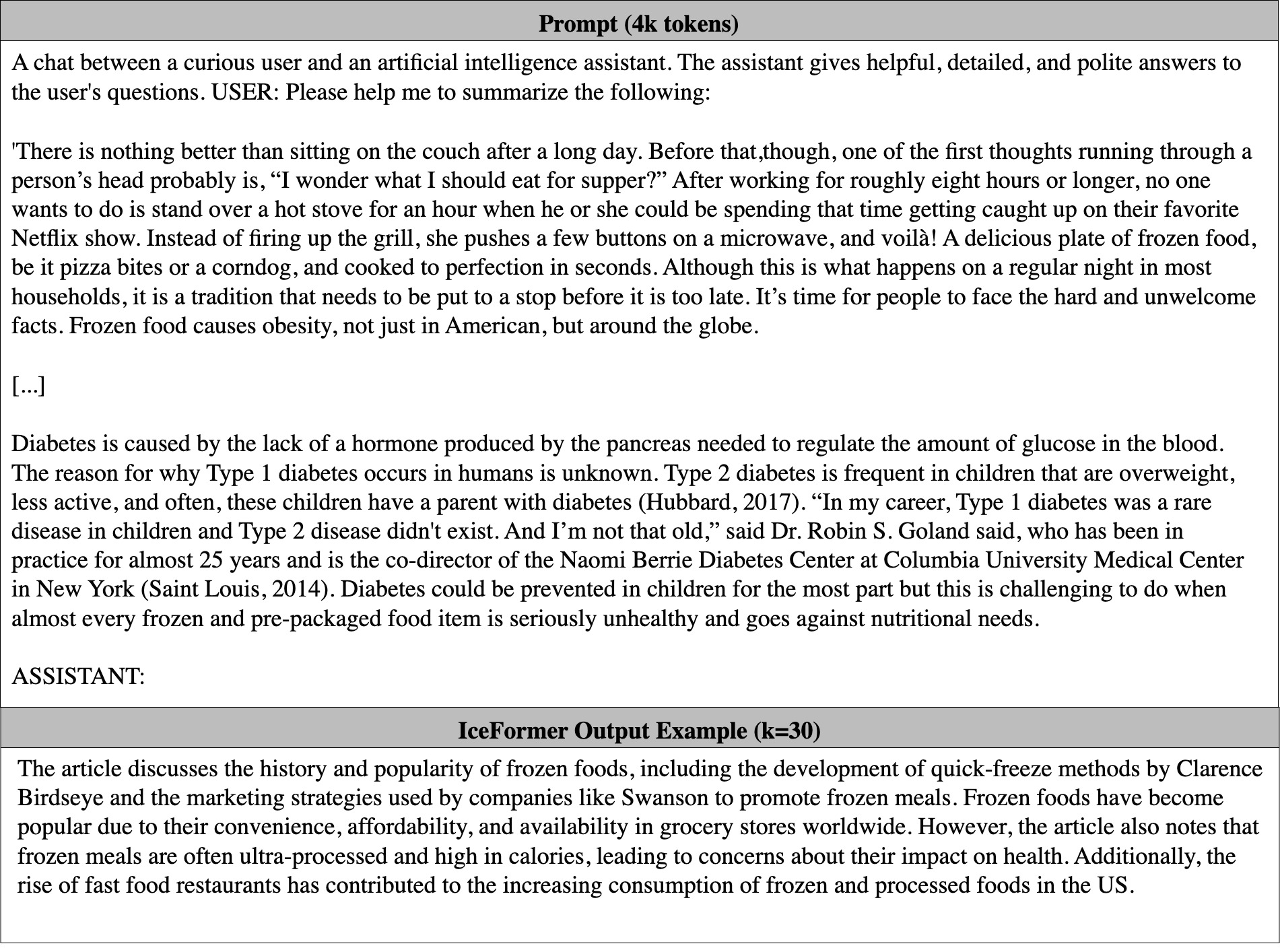}
  \caption{Output of \toolName (Top-30) + LLM with 4k input tokens. We ask the LLM to summarize an article titled ``Frozen Food: The World’s Favorite Killer''.}
  \label{fig:4k}
\end{figure}

\begin{figure}[htb]
  \centering
\includegraphics[width=1.0\linewidth]{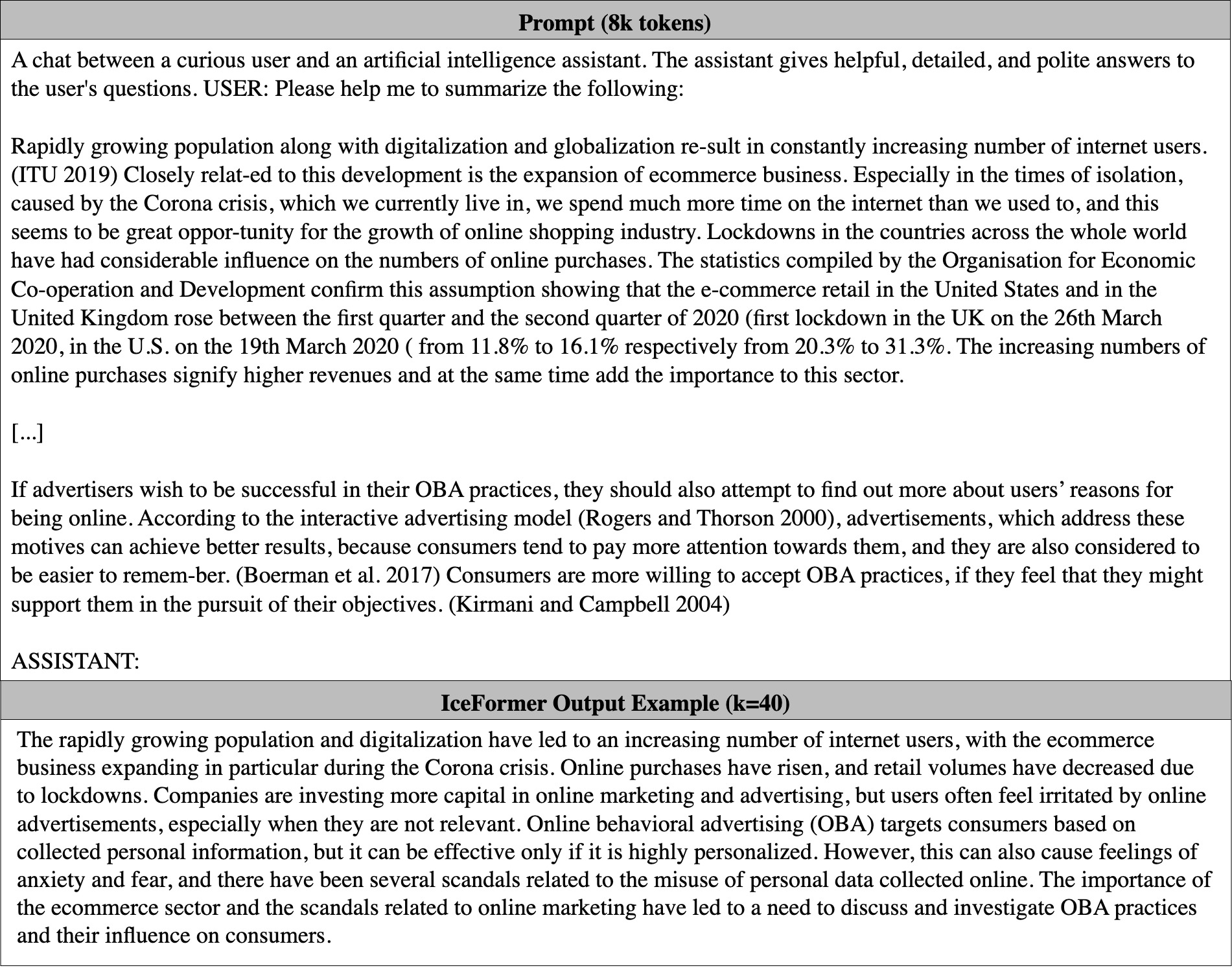}
  \caption{Output of \toolName (Top-40) + LLM with 8k input tokens. We ask the LLM to summarize an article titled ``Research of How Online Behavioral Advertising Influences Consumers''.}
  \label{fig:8k}
\end{figure}

In this section, we provide the text outputs of \toolName when applied to the LLM (vicuna-7b-v1.5-16k) in Figure~\ref{fig:4k}\&\ref{fig:8k}. We also include the full information of the input prompts in our supplementary folder.

% \begin{figure}[htb]
%   \centering
% \includegraphics[width=1.0\linewidth]{short_out.png}
%   \caption{Output of \toolName (Top-20) +  StarCoder with 4.8k input tokens.}
%   \label{fig:short_out}
% \end{figure}

% \begin{figure}[htb]
%   \centering
% \includegraphics[width=1.0\linewidth]{long_out.png}
%   \caption{Output of \toolName (Top-30) +  StarCoder with 6k input tokens.}
%   \label{fig:long_out}
% \end{figure}

% \begin{figure}[htb]
%   \centering
% \includegraphics[width=1.0\linewidth]{MPT_out.png}
%   \caption{Output of \toolName (Top-30) +  MPT-7B-StoryWriter with 12k input tokens.}
%   \label{fig:MPT_out}
% \end{figure}

\end{document}